\documentclass{article}

\usepackage[preprint,nonatbib]{neurips_2021}

\usepackage[utf8]{inputenc} 
\usepackage[T1]{fontenc}    
\usepackage{hyperref}       
\usepackage{url}            
\usepackage{booktabs}       
\usepackage{amsfonts}       
\usepackage{nicefrac}       
\usepackage{microtype}      
\usepackage{xcolor}         
\usepackage{amsmath}
\usepackage{graphicx}
\usepackage{enumitem}
\usepackage{algorithm}
\usepackage{algorithmic}
\usepackage{wrapfig}
\usepackage{subfigure}

\newcommand{\R}{\mathbb{R}}
\newcommand{\I}{\mathbb{I}}
\newcommand{\calL}{\mathcal{L}}
\newcommand{\calC}{\mathcal{C}}
\newcommand{\E}{\mathbb{E}}
\newcommand{\KL}{D_{KL}}
\newcommand{\calN}{\mathcal{N}}

\title{Towards Robust Active Feature Acquisition}

\author{%
  Yang Li \\
  Department of Computer Science\\
  University of North Carolina at Chapel Hill\\
  Chapel Hill, NC 27514 \\
  \texttt{yangli95@cs.unc.edu} \\
   \And
   Siyuan Shan \\
   Department of Computer Science \\
   University of North Carolina at Chapel Hill\\
   Chapel Hill, NC 27514 \\
   \texttt{siyuanshan@cs.unc.edu} \\
   \AND
   Qin Liu \\
   Department of Computer Science \\
   University of North Carolina at Chapel Hill\\
   Chapel Hill, NC 27514 \\
   \texttt{qinliu19@email.unc.edu} \\
   \And
   Junier B. Oliva \\
   Department of Computer Science \\
   University of North Carolina at Chapel Hill\\
   Chapel Hill, NC 27514 \\
   \texttt{joliva@cs.unc.edu} \\
}

\begin{document}

\maketitle

\begin{abstract}
Truly intelligent systems are expected to make critical decisions with incomplete and uncertain data. Active feature acquisition (AFA), where features are sequentially acquired to improve the prediction, is a step towards this goal. However, current AFA models all deal with a small set of candidate features and have difficulty scaling to a large feature space. Moreover, they are ignorant about the valid domains where they can predict confidently, thus they can be vulnerable to out-of-distribution (OOD) inputs. In order to remedy these deficiencies and bring AFA models closer to practical use, we propose several techniques to advance the current AFA approaches. Our framework can easily handle a large number of features using a hierarchical acquisition policy and is more robust to OOD inputs with the help of an OOD detector for partially observed data. Extensive experiments demonstrate the efficacy of our framework over strong baselines.
\end{abstract}

\section{Introduction}
A typical machine learning system will first collect all the features and then predict the target variables based on the collected feature values. Unlike the two-step paradigm, active feature acquisition performs feature value acquisition and target prediction at the same time. Features are actively acquired to improve the prediction, and the prediction in turn informs the acquisition process. Ideally, only features that provide unique information and outweigh their cost will be acquired. The AFA model will stop acquiring more features when the prediction is sufficiently accurate or it exceeds the given acquisition budget. Since each instance could have different set of informative features, active feature acquisition is expected to acquire different features for different instances.

As a motivating example, consider a doctor making a diagnosis on a patient (an instance). The doctor usually has not observed all the possible measurements (such as blood samples, x-rays, etc.) from the patient. The doctor is also not forced to make a diagnosis based on the currently observed measurements; instead, he/she may dynamically decide to take more measurements to help determine the diagnosis. The next measurement to make (feature to observe), if any, will depend on the values of the already observed features; thus, the doctor may determine a different set of features to observe from patient to patient (instance to instance) depending on the values of the features that were observed. Hence, each patient will not have the same subset of features selected (as would be the case with typical feature selection).

In the current literature, there are mainly two types of approaches to acquire features actively: greedy acquisition approaches and reinforcement learning based approaches. Both approaches acquire features sequentially, that is, one candidate feature is acquired at each acquisition step based on the previously observed features. Greedy approaches directly optimize the utility of the next acquisition, while reinforcement learning based approaches optimize a discounted reward along the acquisition trajectories. As a result, the reinforcement learning (RL) based approaches tend to find better solutions to the AFA problem as shown in \cite{li2020active}. Here, we base ourselves on the Markov decision process (MDP) formulation of the AFA problem proposed in \cite{li2020active,shim2018joint} and focus on resolving the deficiencies of the current AFA models.

One of the obstacles to extending the current AFA models to practical use is the potentially large number of candidate features. Greedy approaches are computationally difficult to scale, because the utilities need to be recalculated for each candidate feature based on the updated set of observed features at each acquisition step, which incurs an $O(d^2)$ complexity for a $d$ dimensional feature space. Reinforcement learning algorithms are known to have difficulties with a high dimensional action space \cite{dulac2015deep}. In this work, we propose to cluster the candidate features into groups and use a hierarchical reinforcement learning agent to select the next feature to be acquired.

Another challenge in deploying the AFA models is their robustness. In a practical application, it is very likely for an AFA model to encounter inputs that are different from its training distribution. For example, it may be asked to acquire features for patients with unknown disease. For those out-of-distribution instances, the AFA model may acquire an arbitrary subset of features and predict one of its known categories. Dealing with out-of-distribution inputs is difficult in general, and it is even more challenging for AFA models, since the model only has access to a subset of features at any acquisition step. In this work, we propose a novel algorithm for detecting out-of-distribution inputs with partially observed features, and further utilize it to improve the robustness of the AFA model.

Our contributions are as follows: 1) We propose to reduce the action space for active feature acquisition by grouping similar actions and further learn a hierarchical policy to select the next candidate feature to be acquired. 2) We develop a novel out-of-distribution detection algorithm that can distinguish OOD inputs using an arbitrary subset of features. 3) Armed with the partially observed OOD detection algorithm, we encourage the AFA agent to acquire features that are most informative for distinguishing OOD inputs. 4) Our approach achieves the state-of-the-art performance for active feature acquisition, meanwhile it is more robust to out-of-distribution inputs.

\section{Background}\label{sec:background}
\subsection{Active Feature Acquisition (AFA)}\label{sec:afa}
Typical discriminative models predict a target $y$ using all $d$-dimensional features $x\in\R^d$. AFA, instead, actively acquires feature values to improve the prediction. It typically starts from an empty set of features and sequentially acquires more features $x_{i}$ until the prediction is sufficiently accurate or it exceeds the given acquisition budget. The goal of an AFA model is to minimize the following objective
\begin{equation}
    \calL(\hat{y}(x_{o}),y) + \alpha \calC(o),
\label{eq:objetive}
\end{equation}
where $\calL(\hat{y}(x_{o}),y)$ is the prediction error between the groundtruth target $y$ and the prediction $\hat{y}$ using the acquired features $x_o$, $\calC(o)$ measures the total cost of acquiring a subset of features $o \subseteq \{1,\ldots,d\}$, and $\alpha$ balances these two terms.

However, directly optimizing \eqref{eq:objetive} is not trivial, since it involves optimizing over a combinatorial number of possible subsets. Many heuristic approaches have been developed to approximately solve this problem. For example, in \cite{ling2004decision}, the authors propose to take into account the cost of features when selecting an attribute for building a decision tree so that the final tree will have a minimum total cost. \cite{chai2004test} utilizes a naive Bayes classifier to handle the partially observed features, where the unobserved features are simply ignored in the likelihood objective. They then assess the utility of each unobserved feature by their expected reduction of the misclassification cost. At each acquisition step, the feature with highest utility is acquired. Authors in \cite{nan2014fast} instead leverage a margin-based classifier. An instance is classified by retrieving the nearest neighbors from training set using the partially observed features, and the utility of each unobserved feature is calculated by the one-step ahead classification accuracy. Following the same greedy solution, EDDI \cite{ma2018eddi} utilizes the modern generative models to handle partially observed instances. Specifically, they propose a partial VAE to model the arbitrary marginal likelihoods $p(x_o)$ (target variable $y$ is concatenated into $x$ and modeled together). Inspired by the experimental design approaches \cite{bernardo1979expected}, they assess the utility of each unobserved feature with their expected information gain to the target variable $y$, i.e.,
\begin{equation}
    \mathcal{U}_i = \E_{p(x_i \mid x_o)} \KL[p(y \mid x_o, x_i) \| p(y \mid x_o)].
\end{equation}
The feature with highest utility is acquired at each step. Similar to EDDI, Icebreaker \cite{gong2019icebreaker} proposes to use a Bayesian Deep Latent Gaussian model to capture the uncertainty of unobserved features and to assess their utilities. They further extend the problem to actively acquire additional information during training.

Greedy approaches are easy to understand, but they are also inherently flawed, since they are myopic and unaware of the long-term goal of obtaining multiple features that are \emph{jointly} informative. Instead of acquiring features greedily, the AFA problem has been formulated as a Markov Decision Process (MDP) \cite{zubek2004pruning,ruckstiess2011sequential}. Therefore, reinforcement learning based approaches can be utilized, where a long-term discounted reward is optimized. In the MDP formulation of AFA problem, the state is the current observed features, the action is the next feature to acquire, and the reward contains the final prediction reward and the cost of each acquired feature. In \cite{li2020active}  and \cite{shim2018joint}, a special action indicating the termination of the acquisition process is also introduced. The agent will stop acquiring more features when it selects the termination action. Specifically, we have
\begin{equation}
    s = [o, x_o], \qquad a \in u \cup \phi, \qquad r(s,a)=-\calL(\hat{y}(x_{o}),y)\I(a=\phi)-\alpha \calC(a)\I(a \neq \phi),
\label{eq:mdp}
\end{equation}
where the state, $s$, consists of the current acquired feature subset, $o \subseteq \{1,\ldots,d\}$, and their values, $x_o$. The action, $a$, is either one of the remaining unobserved features, $u=\{1, \ldots, d\} \setminus o$, or the termination action, $\phi$. When a new feature, $i$, is acquired, the current state transits to a new state following $o \xrightarrow{i} o \cup i$, $x_o \xrightarrow{i} x_o \cup x_i$, and the agent receives the negative acquisition cost of this feature as a reward. If the termination action is selected (i.e., $a = \phi$), the agent makes a prediction based on all acquired features, $x_o$, and receives a final reward as $-\calL(\hat{y}(x_{o}),y)$.

Given the above MDP formulation, several RL approaches have been explored. \cite{zubek2004pruning} fits a transition model using complete data, and then uses the $\text{AO}^*$ heuristic search algorithm to find an optimal policy. \cite{ruckstiess2011sequential} utilizes Fitted Q-Iteration to optimize the MDP. \cite{he2012imitation} and \cite{he2016active} instead employ a imitation learning approach coached by a greedy reference policy. JAFA \cite{shim2018joint} jointly learns an RL agent and a classifier, where the classifier is deemed as the environment to calculate the reward. 

Although MDPs are broad enough to encapsulate the active acquisition of features, there are several challenges that limit the success of a naive reinforcement learning approach. In the aforementioned MDP, the agent pays the acquisition cost at each acquisition step but only receives a reward about the prediction after completing the acquisition process. This results in sparse rewards leading to credit assignment problems for potentially long episodes \cite{minsky1961steps,sutton1988learning}, which may make training difficult. In addition, an agent that is making feature acquisitions must also navigate a complicated high-dimensional action space, as the action space scales with the number of features, making for a challenging RL problem \cite{dulac2015deep}. Finally, the agent must manage multiple roles as it has to: implicitly model dependencies (to avoid selecting redundant features); perform a cost/benefit analysis (to judge what unobserved informative feature is worth the cost); and act as a classifier (to make a final prediction). Attaining these roles individually is a challenge, doing so jointly and without gradients on the reward (as with the MDP) is an even more daunting problem. To lessen the burden on the agent, and assuage the other challenges, GSMRL \cite{li2020acflow} proposes a model-based alternative. The key observation of GSMRL is that the dynamics of the above MDP can be modeled by the conditional dependencies among features.
That is, the state transitions are based on the conditionals: $p(x_j \mid x_o)$, where $x_o$ is current observed features and $x_j$ is an unobserved feature. Base on this observation, a surrogate model that captures the arbitrary conditionals, $p(x_u, y \mid x_o)$, is employed to assist the agent. 
Specifically, GSMRL defines an intermediate reward using the information gain of the acquired feature, $x_i$, at the current acquisition step, i.e.,
$
    r_m(s,i) = H(y \mid x_o) -  H(y \mid x_o, x_i).
$
The entropy terms are estimated using the learned surrogate model.
Furthermore, the expected information gain for each candidate acquisition is provided to the agent as side information, i.e.,
\begin{equation}
\mathcal{U}_j = H(y \mid x_o) - \E_{p(x_j \mid x_o)}H(y \mid x_o, x_j), \quad j \in u.
\end{equation}
In addition to $\mathcal{U}_j$, the surrogate model can also provide the current prediction $\hat{y}$, the prediction probability, $p(y \mid x_o)$, the imputed values of unobserved features and their uncertainties, $p(x_u \mid x_o)$, as auxiliary information. Armed with the auxiliary information and the intermediate reward, GSMRL alleviates the challenge of a model-free approach and obtains state-of-the-art performance for several AFA problems. Given their established excellence, we use GSMRL as the base model for our robust AFA framework.

\subsection{Active Instance Recognition (AIR)}
AFA acquires features actively to improve the prediction of a target variable, while some application do not have an explicit target; instead, the features are acquired to improve our understanding to the instance. In GSMRL \cite{li2020active}, the authors propose a task named AIR, where an agent acquires features actively to reconstruct the unobserved features. A similar model-based RL approach is used for AIR, where a dynamics model $p(x_u \mid x_o)$ captures the state transition. The intermediate reward and auxiliary information can be similarly derived by replacing $y$ with $x_{u \setminus i}$ (the unobserved features excluding the current candidate $i$). A special case of AIR is to acquire features in k-space for accelerated MRI reconstruction. \cite{pineda2020active} and \cite{bakker2020experimental} have explored this application using deep Q-learning and policy gradient respectively. Several non-RL approaches \cite{zhang2019reducing,gorp2021active} have also been proposed.

\subsection{Out-of-distribution Detection}\label{sec:ood}
ML models are typically trained with a specific data distribution, however, when deployed, the model may encounter data that is outside the training distribution. For those out-of-distribution inputs, the prediction could be arbitrarily bad. Therefore, detecting OOD inputs has been an active research direction. One approach relies on the uncertainty of prediction. The prediction for OOD inputs are expected to have higher uncertainty. However, deep classifiers tend to be overly confident about their prediction, thus a postprocessing step to calibrate the uncertainty is needed \cite{ovadia2019can,kumar2019verified}. Bayesian neural networks (BNNs), where weights are sampled from their posterior distributions, have also been used to quantify the uncertainty \cite{blundell2015weight}. However, the full posterior is usually intractable. To avoid the complex posterior inference of BNNs, deep ensemble \cite{lakshminarayanan2016simple} and MC dropout \cite{gal2016dropout} are proposed as two approximations, where multiple independent forward passes are conducted to obtain a set of predictions. To avoid the expensive multiple forward passes, DUQ \cite{van2020uncertainty} propose to detect OOD inputs with a single deterministic model in a single pass. It learns a set of centroid vectors corresponding to the different classes and measures uncertainty according to the distance calculated by an RBF network \cite{lecun1998gradient} between the model output and its closest centroid. Recently, Gaussian processes have been combined with deep neural networks to quantify the prediction uncertainty, such as SNGP \cite{liu2020simple} and DUE \cite{van2021improving}.

Intuitively, generative models are expected to be able to detect OOD inputs using the likelihood \cite{bishop1994novelty}, that is, OOD inputs should have lower likelihood than the in-distribution ones. However, recent works \cite{nalisnick2018do, hendrycks2018deep} show that it is not the case for high dimensional distributions. This oddity has been verified for many modern deep generative models, such as PixelCNN, VAE and normalizing flows. To rectify this pathology, \cite{choi2018waic} proposes to train multiple generative models in parallel and utilize the Watanabe-Akaike Information Criterion to identify OOD data. \cite{ren2019likelihood} proposes to use likelihood ratio of an autoregressive model to remove the contributions from background pixels. \cite{nalisnick2019detecting} propose a simple typicality test for a batch of data.
Furthermore, \cite{morningstar2021density} propose to utilize an estimator for the density of states on several summary statistics of the in-distribution data. It then evaluates the density of states estimator (DoSE) on testing data and marks those with low support as OOD. MSMA \cite{mahmood2021multiscale} empirically observes that the norm of scores conditioned on different noise levels serves as very effective summary statistics in the DoSE framework, and thus utilizes the multi-scale score matching network \cite{song2019generative} to calculate those statistics.

\section{Method}
In this section, we introduce each component of our framework. We use the model-based AFA approach, GSMRL \cite{li2020active}, as the base model, which utilizes an arbitrary conditional model $p(x_u, y \mid x_o)$ to assist the agent by providing the intermediate rewards and the auxiliary information. We further leverage the arbitrary conditionals to cluster features into groups and develop a hierarchical acquisition policy to deal with the large action space. Next, we introduce the OOD detection algorithm for partially observed instances along the acquisition trajectories. We then compose all those components together and propose the robust active feature acquisition framework. For convenience of evaluating OOD detection performance, we do not use the termination action for AFA but specify a budget of the acquisition (i.e., the number of features being acquired).

\subsection{Action Space Grouping}
As described in Sec.~\ref{sec:afa}, the AFA problem can be interpreted as a MDP, where the action space at each acquisition step contains the current unobserved features. For certain problems, the action space could be enormous. For example, in the aforementioned health care example, the action space could contain an exhaustive list of possible inspections a hospital can offer. Dealing with large action space for RL is generally challenging, since the agent may not be able to effectively explore the entire action space. Several approaches have been proposed to train RL agent with a large discrete action space. For instance, \cite{dulac2015deep} proposes a Wolpertinger policy that maps a state representation to a continuous action representation. The action representation is then used to look up $k$-nearest valid actions as candidates. Finally, the action with the highest Q value is selected and executed in the environment. \cite{majeed2020exact} proposes a sequentialization scheme, where the action space is transformed into a sequence of $\mathcal{B}$-ary decision code words. A pair of bijective encoder-decoder is defined to perform this transformation. Running the agent will produce a sequence of decisions, which are subsequently decoded as a valid action that can be executed in the environment.

Similar to \cite{majeed2020exact}, we also formulate the action space as a sequence of decisions. Here, we propose to utilize the inherited clustering properties of the candidate features. Given a set of features, $\{x_1,\ldots,x_d\}$, we assume features can be clustered based on their informativeness to the target variable $y$. That is, there might be a subset of features that are decisive about $y$ and another subset of features that are not relevant to $y$. This assumption holds true for many real-life tasks. For example, a music recommender system might use a questionnaire to collect information from a user. Some questions about musicians or song genres are closely related to the target, while some questions about addresses might not be relevant at all. Based on this intuition, we propose to assess the informativeness of the candidate features using their mutual information to the target variable, $y$, i.e., $I(x_i ; y)$, where $i \in \{1,\ldots,d\}$. The mutual information can be estimated using the learned arbitrary conditionals of the surrogate model
\begin{equation}
    I(x_i; y) = \E_{x_i,y} \log \frac{p(x_i, y)}{p(x_i)p(y)} = \E_{x_i,y} \log \frac{p(y \mid x_i)}{p(y \mid \emptyset)},
\end{equation}
where the expectation is estimated using a held-out validation set. Given the estimated mutual information, we can simply sort and divide the candidate features into different groups. For the sake of implementation simplicity, we use clusters with the same number of features. We can further group features inside each cluster into smaller clusters and develop a tree-structured action space as in \cite{majeed2020exact}, which we leave for future works. Note that the clustering is not performed actively for each instance; instead, we cluster once for each dataset and keep the cluster structure fixed throughout the acquisition process.

It is worth noting that the mutual information $I(x_i ; y)$ is not the only choice for clustering features. Alternative quantities, such as the mutual information $I(x_i ; x_j)$ or a metric $d(x_i, x_j) = H(x_i, x_j) - I(x_i ; x_j)$, can be used together with a hierarchical clustering procedure to group candidate features. However, these alternatives need to be estimated for each pair of candidate features, which incurs a $O(d^2)$ computational complexity, while the mutual information, $I(x_i ; y)$, only has $O(d)$ complexity.

\begin{figure}
\begin{minipage}{0.47\linewidth}
    \centering
    \includegraphics[width=0.87\linewidth]{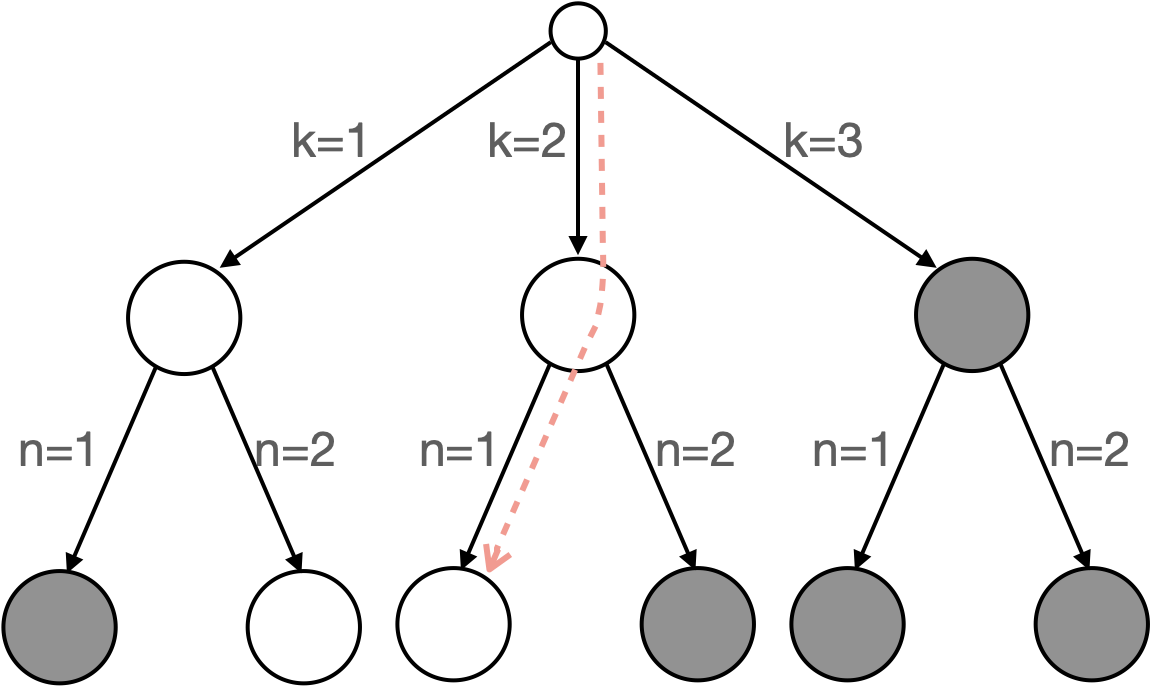}
    \caption{An illustrative example of the grouped action space, where 6 features are grouped into 3 clusters. The grayed circles represent the current observed features (or fully observed groups) and are not considered as candidates anymore. The dashed line shows one acquisition at the current step, which acquires the feature $g_2^{(1)}$. The corresponding circles will be grayed after this acquisition step.}
    \label{fig:action_space}
\end{minipage}
\hfill
\begin{minipage}{0.48\linewidth}
\begin{algorithm}[H]
\caption{Robust Active Feature Acquisition}
\label{alg:robust_afa}
\begin{algorithmic}
\REQUIRE{acquisition environment $\textit{env}$; dynamics model $\textit{M}$; partially observed OOD detector $\textit{D}$; AFA agent $\textit{agent}$; acquisition budget $\textit{B}$}
\ENSURE{reward, ood\_likelihood, prediction}
\STATE{$x_o$, o, reward = \textit{env}.reset()}
\WHILE{$|o| < \textit{B}$}{
  \STATE{aux = $\textit{M}$.query($x_o$, o)}
  \STATE{action = \textit{agent}.act($x_o$, o, aux)}
  \STATE{$r_m =$ $\textit{M}$.reward($x_o$, $o$, action)}
  \STATE{$x_o$, o, $r_e$ = \textit{env}.step(action)}
  \STATE{reward += $r_e$ + $r_m$}}
\ENDWHILE
\STATE{aux = $\textit{M}$.query($x_o$, o)}
\STATE{prediction = $\textit{agent}$.predict($x_o$, o, aux)}
\STATE{$r_p = $ $\textit{env}$.reward($x_o$, $o$, prediction)}
\STATE{$r_d =$ $\textit{D}$.reward($x_o$, $o$)}
\STATE{reward += $r_p$ + $r_d$}
\STATE{ood\_likelihood = $\textit{D}$.log\_prob($x_o$, o)}
\end{algorithmic}
\end{algorithm}
\end{minipage}
\end{figure}

Given the grouped action space, $\mathcal{A} = \{g_k\}_{k=1}^{K}$, with $K$ distinct clusters, we develop a hierarchical policy to select one candidate feature at each acquisition step. $g_k = \{g_k^{(1)}, \ldots, g_k^{(N)}\} \subseteq \{1,\ldots,d\}$ represents the $k_{th}$ group of features of size $N$, where $\forall k \neq k',\ g_k \cap g_{k'} = \emptyset$ and $\cup_{k=1}^K g_k = \{1, \ldots, d\}$. The policy factorizes autoregressively by first selecting the group index, $k$, and then selecting the feature index, $n$, inside the selected group, i.e.,
\begin{equation}\label{eq:autoreg_action}
    p(a \mid s) = p(k \mid s)p(n \mid k, s), \quad k \in \{1, \ldots, K\}, \quad n \in \{1, \ldots, N\}.
\end{equation}
The actual feature index being acquired is then decoded as $g_k^{(n)}$. As the agent acquires features, the already acquired features are removed from the candidate set. We simply set the probabilities of those features to zeros and renormalize the distribution. Similarly, if all features of a group have been acquired, the probability of this group is set to zero. With the proposed action space grouping, the original $d$-dimensional action space is reduced to $K+N$ decisions. Please refer to Fig.~\ref{fig:action_space} for an illustration.

\subsection{Partially Observed Out-of-distribution Detection}\label{sec:po3d}
In Sec.~\ref{sec:ood}, we introduce several advanced techniques to detect out-of-distribution inputs. However, those approaches require fully observed data. In an AFA framework, data are partially observed at any acquisition step, which renders those approaches inappropriate. In this section, we develop a novel OOD detection algorithm specifically tailored for partially observed data. 
Inspired by MSMA \cite{mahmood2021multiscale}, we propose to use the norm of scores from an arbitrary marginal distribution $p(x_o)$ as summary statistics and further detect partially observed OOD inputs with a DoSE \cite{morningstar2021density} approach.
MSMA for fully observed data is built by the following steps:
\begin{enumerate}[label=(\roman*)]
    \item Train a noise conditioned score matching network $s_\theta$ \cite{song2019generative} with $L$ noise levels by optimizing
    \begin{equation}\label{eq:score_matching}
        \frac{1}{L}\sum_{i=1}^{L} \frac{\sigma_i^2}{2} \E_{p_{data}(x)}\E_{\tilde{x} \sim \calN(x, \sigma_i^2 I)} \left[ \left\Vert s_\theta(\tilde{x}, \sigma_i) + \frac{\tilde{x}-x}{\sigma_i^2} \right\Vert_2^2 \right].
    \end{equation}
    The score network essentially approximates the score of a series of smoothed data distributions $\nabla_{\tilde{x}} \log q_{\sigma_i}(\tilde{x})$, where $q_{\sigma_i}(\tilde{x}) = \int p_{data}(x)q_{\sigma_i}(\tilde{x} \mid x) dx$, and $q_{\sigma_i}(\tilde{x} \mid x)$ transforms $x$ by adding some Gaussian noise form $\calN(0, \sigma_i^2I)$.
    
    \item For a given input $x$, compute the L2 norm of scores at each noise level, i.e., $s_i = \Vert s_\theta(x, \sigma_i) \Vert$.
    
    \item Fit a low dimensional likelihood model for the norm of scores using in-distribution data, i.e., $p(s_1, \ldots, s_L)$, which is called density of states in \cite{morningstar2021density} following the concept in statistical mechanics.
    
    \item Threshold the likelihood to determine whether the input $x$ is OOD or not.
\end{enumerate}
In order to deal with partially observed data, we modify the score network to output scores of arbitrary marginal distributions, i.e., $\nabla_{\tilde{x}_m}\log q_{\sigma_i}(\tilde{x}_m)$, where $m \subseteq \{1, \ldots, d\}$ represents an arbitrary subset of features. The training objective \eqref{eq:score_matching} is modified accordingly to
\begin{equation}
    \frac{1}{L}\sum_{i=1}^{L} \frac{\sigma_i^2}{2} \E_{p_{data}(x)}\E_{\tilde{x} \sim \calN(x, \sigma_i^2 I)}\E_{m \sim p(m)} \left[ \left\Vert s_\theta(\tilde{x} \odot \I_m, \I_m, \sigma_i) \odot \I_m + \frac{\tilde{x} \odot \I_m-x \odot \I_m}{\sigma_i^2} \right\Vert_2^2 \right],
\end{equation}
where $\I_m$ represents a $d$-dimensional binary mask indicating the partially observed features, $\odot$ represents the element-wise product operation, and $p(m)$ is the distribution for generating observed dimensions. Similar to the fully observed case, we compute the L2 norm of scores at each noise level, i.e., $s_i = \Vert s_\theta(x \odot \I_m, \I_m, \sigma_i) \odot \I_m \Vert$, and fit a likelihood model in this transformed low-dimensional space. The likelihood model is also conditioned on the binary mask $\I_m$ to indicate the observed dimensions, i.e., $p(s_1,\ldots,s_L \mid \I_m)$. Given an input $x$ with observed dimensions $m$, we threshold the likelihood $p(s_i, \ldots, s_L \mid \I_m)$ to determine whether the partially observed data $x_m$ is OOD or not. 
To train the partially observed MSMA (PO-MSMA), we generate a mask for each input data $x$ at random. The conditional likelihood over norm of scores is estimated by a conditional autoregressive model, for which we utilize the efficient masked autoregressive implementation \cite{papamakarios2017masked}.

One benefit of our proposed PO-MSMA approach is that a single model can be used to detect OOD inputs with arbitrary observed features, which is convenient for detecting OOD inputs along the acquisition trajectories. Furthermore, sharing weights across different tasks (i.e., different marginal distributions) could act as a regularization (as discussed in \cite{li2020acflow}), thus the unified score matching network can potentially perform better than separately trained ones for each different conditional, which we will investigate in future works.

\vspace{-5pt}
\subsection{Robust Active Feature Acquisition}\label{sec:robust_afa}
\vspace{-2pt}
Above, we introduce our proposed action space grouping technique and a partially observed OOD detection algorithm. Combining those components, we can now actively acquire features for problem with a large action space and simultaneously detect OOD inputs using the acquired subset of features. However, the OOD detection performance might be suboptimal, since the agent is not informed of the detection goal and merely focuses on predicting the target variable $y$. The features that are informative for predicting the target might not be informative for distinguishing OOD inputs.

In order to guide the agent to acquire features that are informative for OOD detection, we propose to utilize the likelihood, $p(s_1,\ldots,s_L \mid \I_m)$, as an auxiliary reward, which encourages the agent to acquire features more closely resemble the in-distribution ones, and thus reduces the false positive detection. Alternatively, if reducing the false negative is desired, we could use the negative likelihood $-p(s_1, \ldots, s_L \mid \I_m)$ as the auxiliary reward.

\begin{wrapfigure}{r}{0.35\linewidth}
\centering
\vspace{-10pt}
\includegraphics[width=\linewidth]{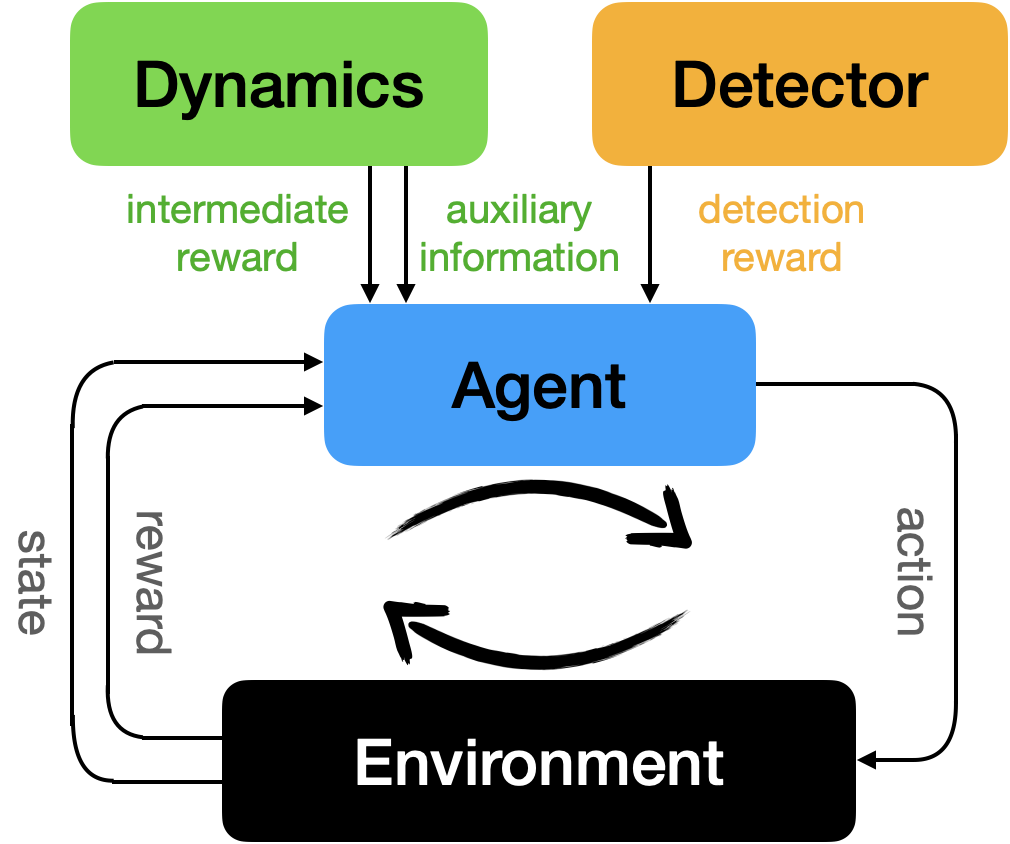}
\vspace{-10pt}
\caption{Schematic illustration of our robust AFA framework.}
\label{fig:robust_afa}
\end{wrapfigure}

In summary, our robust AFA framework contains a dynamics model, an OOD detector and an RL agent. The dynamics model captures the arbitrary conditionals, $p(x_u, y \mid x_o)$, and is utilized to provide auxiliary information and intermediate rewards. It also enables a simple and efficient action space grouping technique and thus scales AFA up to applications with large action spaces. The partially observed OOD detector is used to distinguish OOD inputs alongside the acquisition procedure and also used to provide an auxiliary reward so that the agent is encouraged to acquire informative features for OOD detection. The RL agent takes in the current acquired features and auxiliary information from the dynamics model and predicts what next feature to acquire. When the feature is actually acquired, the agent pays the acquisition cost of the feature and receives an intermediate reward from the dynamics model. When the acquisition process is terminated, the agent makes a final prediction about the target, $y$, using all its acquired features and receives an reward about its prediction. It also receives an reward from the OOD detector about the likelihood of the acquired feature subset in the transformed space (i.e., the norm of the scores). Please refer to Algorithm~\ref{alg:robust_afa} for additional details and to Fig.~\ref{fig:robust_afa} for an illustration.

In GSMRL \cite{li2020active}, the acquisition procedure is terminated when the agent selects a special termination action, which means each instance could have different number of features acquired. Although intriguing for practical use, it introduces additional complexity to assess OOD detection performance, since we need to separate two possible causes of a detection failure, i.e., not sufficient acquisitions and not effective acquisitions. To simplify the evaluation, we instead specify a fixed acquisition budget (i.e., the number of acquired features). The agent will terminate the acquisition process when it exceeds the specified acquisition budget. However, it is possible to incorporate a termination action into our framework.

\vspace{-5pt}
\section{Experiments}
\vspace{-5pt}
In this section, we evaluate our framework on several commonly used OOD detection benchmarks. Our model actively acquires features to predict the target and meanwhile determines whether the input is OOD using only the acquired features. Given that these benchmarks typically have a large number of candidate features, current AFA approaches cannot be applied directly. We instead compare to a modified GSMRL algorithm, where candidate features are clustered with our proposed action space grouping technique. We also compare to a simple random acquisition baseline, where a random unobserved feature is acquired at each acquisition step. The random policy is repeated for 5 times and the metrics are averaged from different runs. Please refer to Appendix~\ref{sec:appendix_exp} for experimental details. For each dataset, we assess the performance under several prespecified acquisition budgets. For classification task, the performance is evaluated by the classification accuracy; for reconstruction task, the performance is evaluated by the reconstruction MSE. We also detect OOD inputs using the acquired features and report the AUROC scores.

\begin{figure}
    \centering
    \begin{minipage}{\linewidth}
    \subfigure[MNIST]{\includegraphics[width=0.3\linewidth]{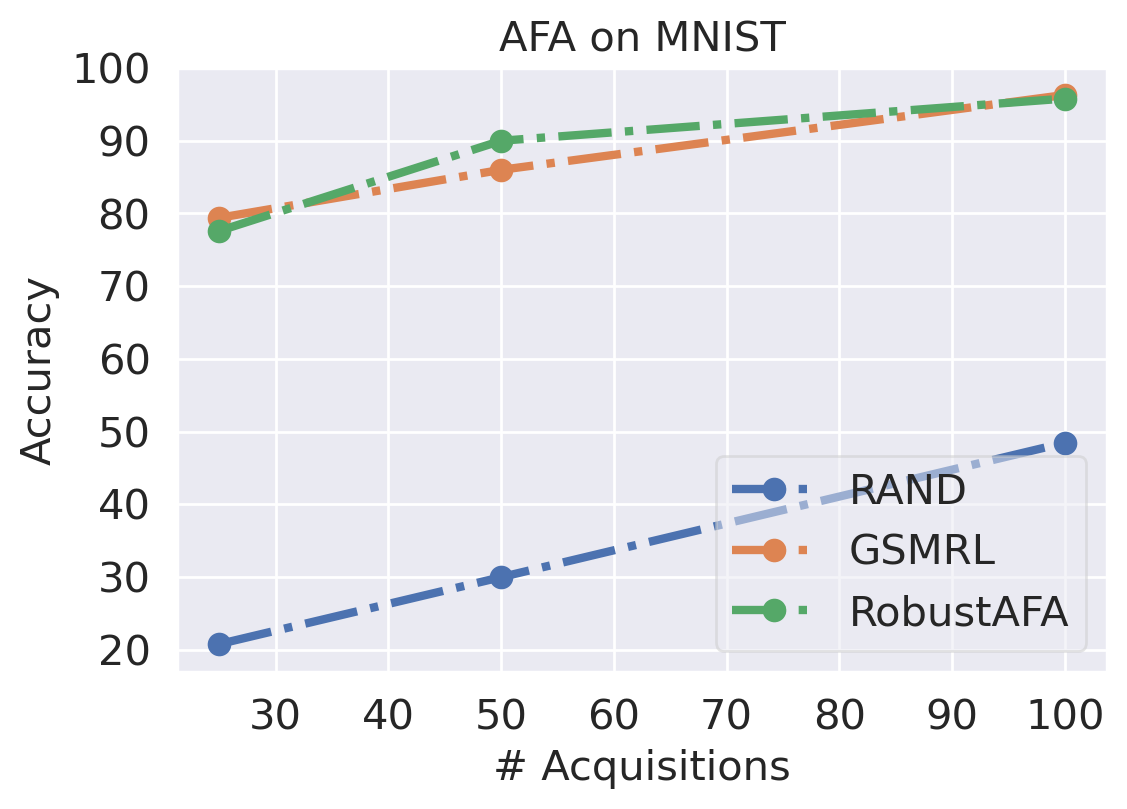}}
    \hspace{3pt}
    \subfigure[FMNIST]{\includegraphics[width=0.3\linewidth]{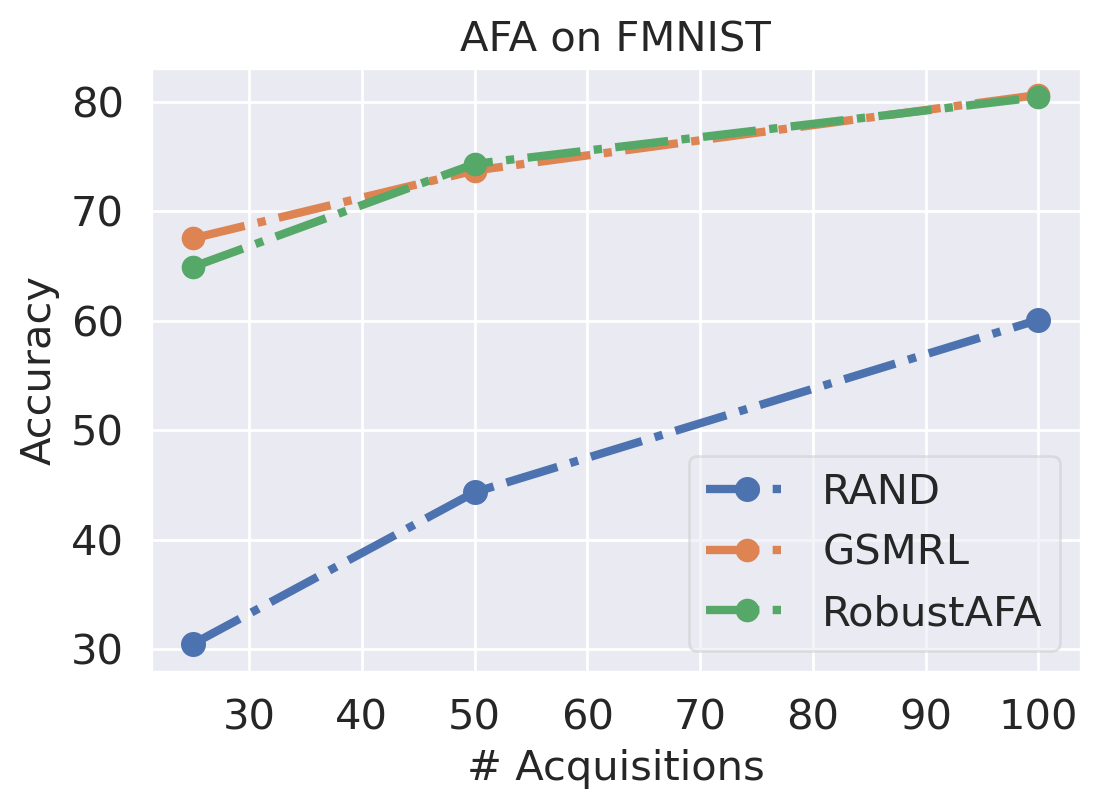}}
    \hspace{3pt}
    \subfigure[SVHN]{\includegraphics[width=0.3\linewidth]{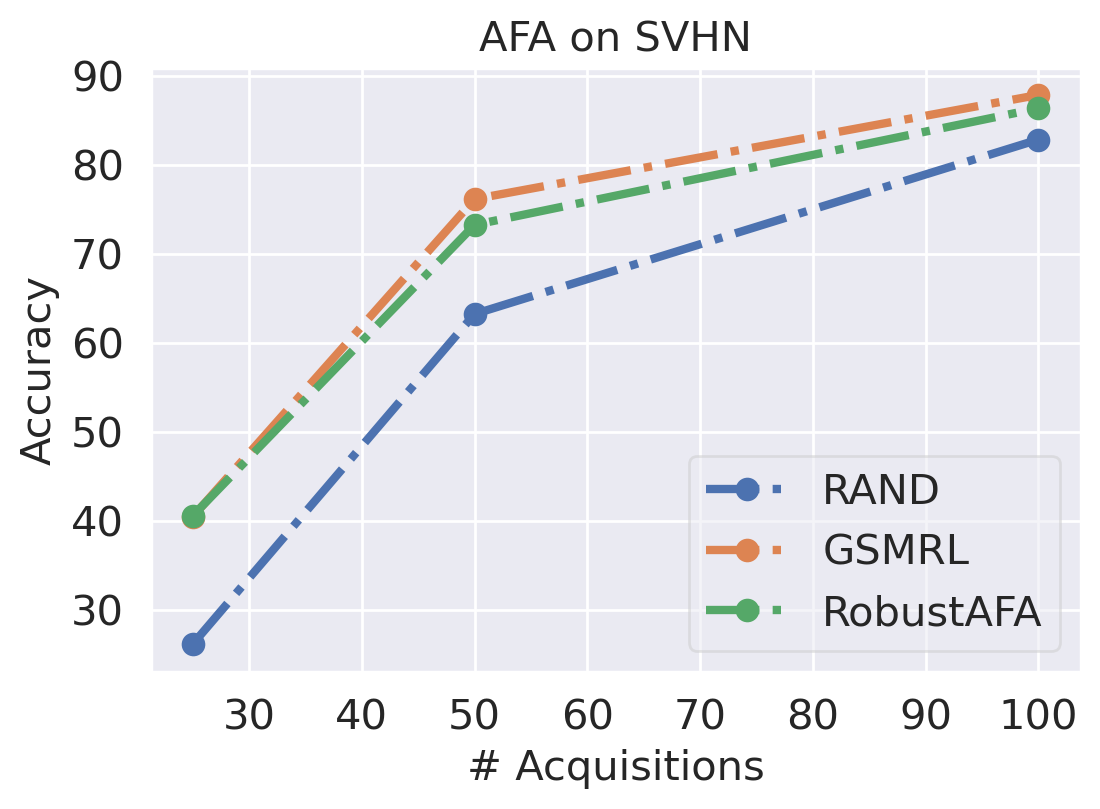}}
    \vspace{-5pt}
    \caption{Classification accuracy for acquiring different number of features.}
    \label{fig:acc}
    \end{minipage}
    \begin{minipage}{\linewidth}
    \subfigure[MNIST - Omniglot]{\includegraphics[width=0.3\linewidth]{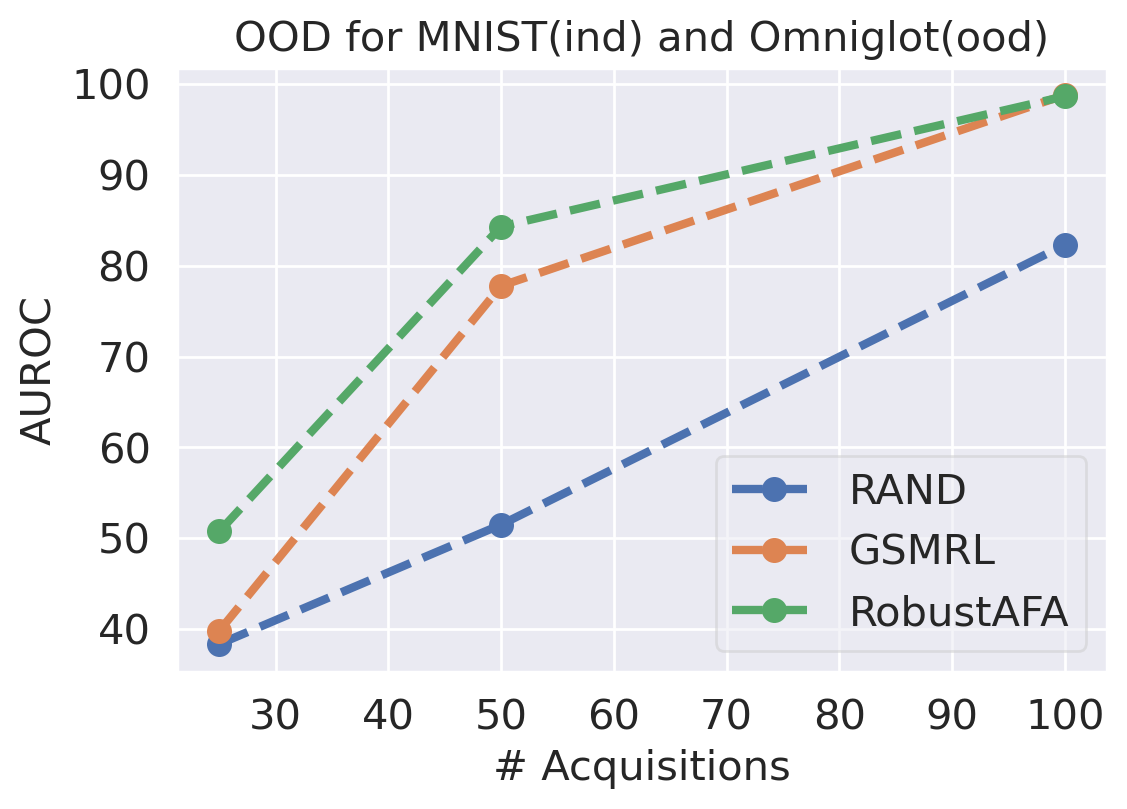}}
    \hspace{3pt}
    \subfigure[FMNIST - MNIST]{\includegraphics[width=0.3\linewidth]{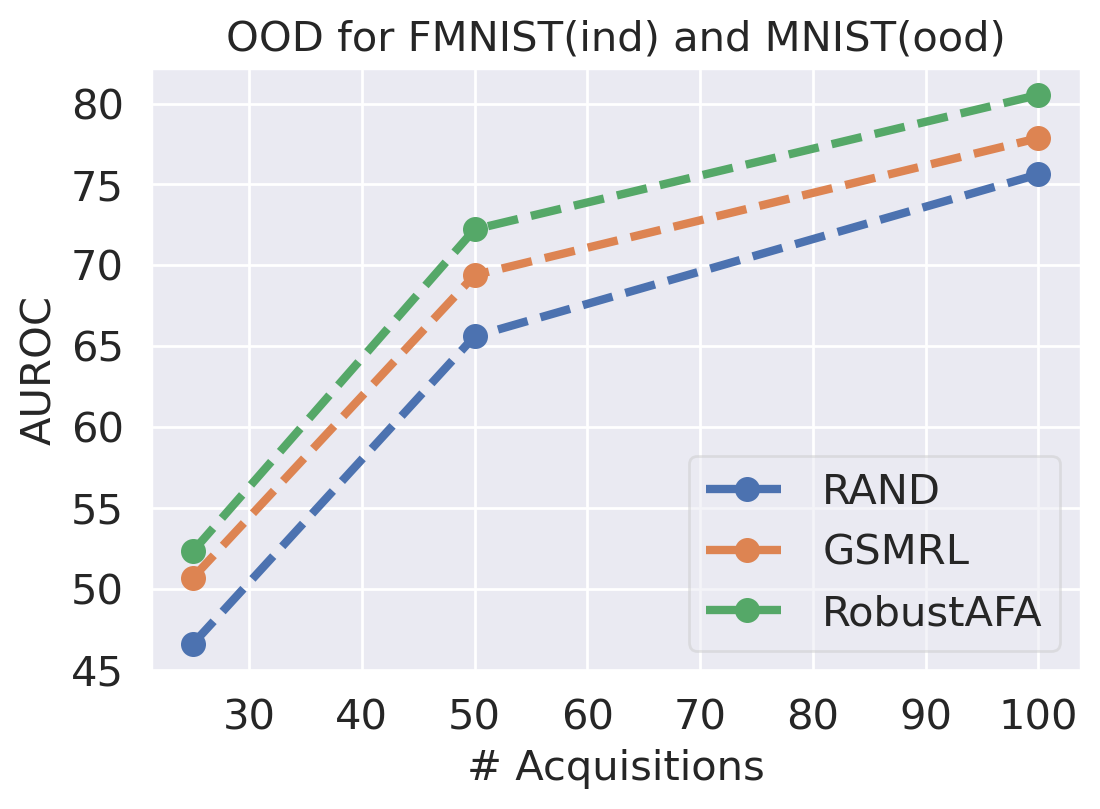}}
    \hspace{3pt}
    \subfigure[SVHN - CIFAR10]{\includegraphics[width=0.3\linewidth]{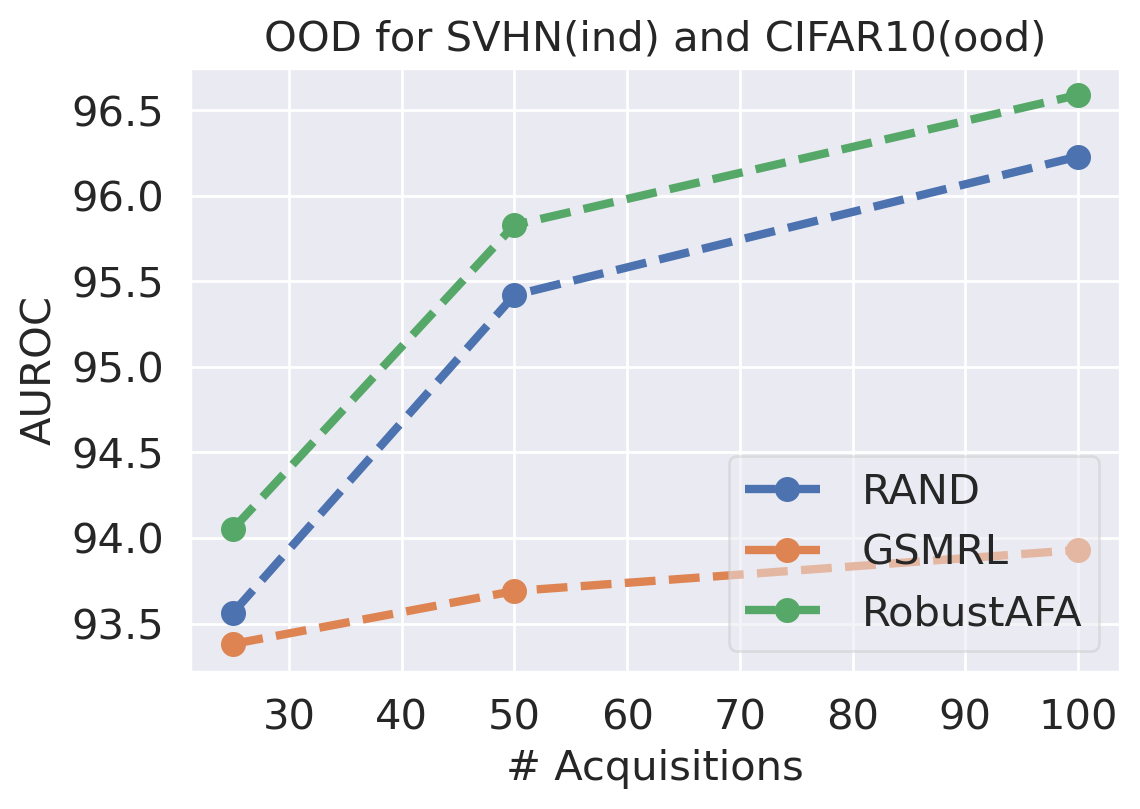}}
    \vspace{-5pt}
    \caption{AUROC for OOD detection with acquired features.}
    \label{fig:auc}
    \end{minipage}
    \vspace{-12pt}
\end{figure}

\begin{wrapfigure}{r}{0.5\linewidth}
    \centering
    \vspace{-15pt}
    \includegraphics[width=\linewidth]{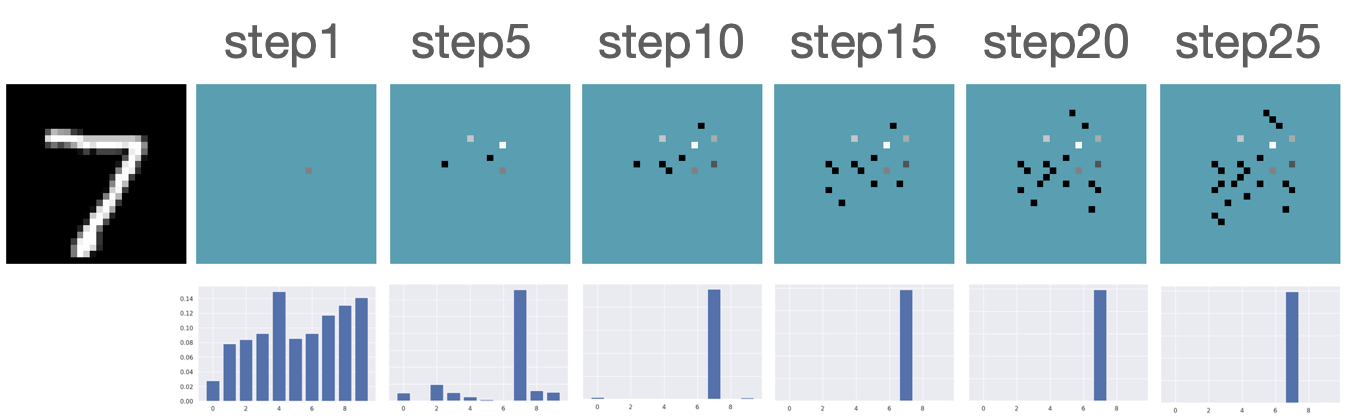}
    
    \vspace{5pt}
    
    \includegraphics[width=\linewidth]{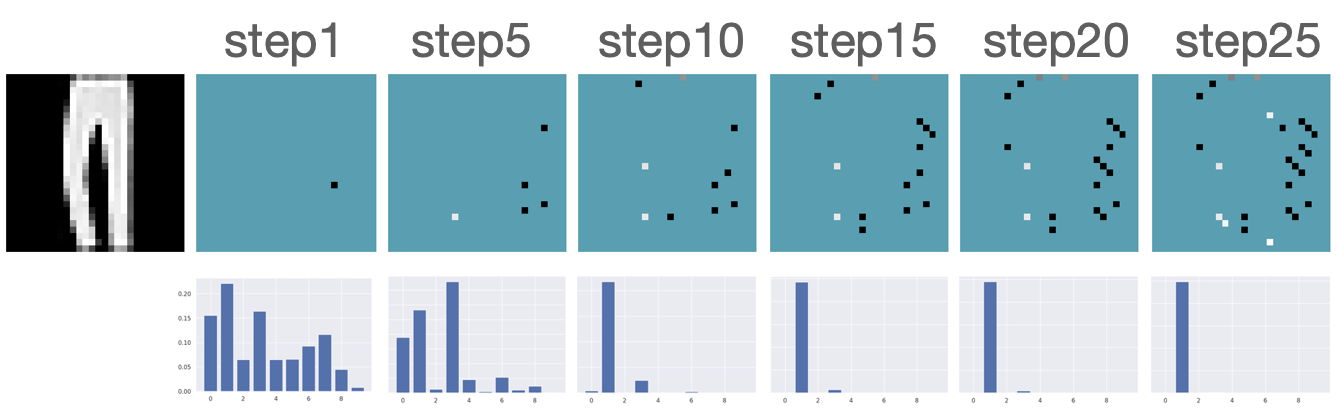}
    
    \vspace{5pt}
    
    \includegraphics[width=\linewidth]{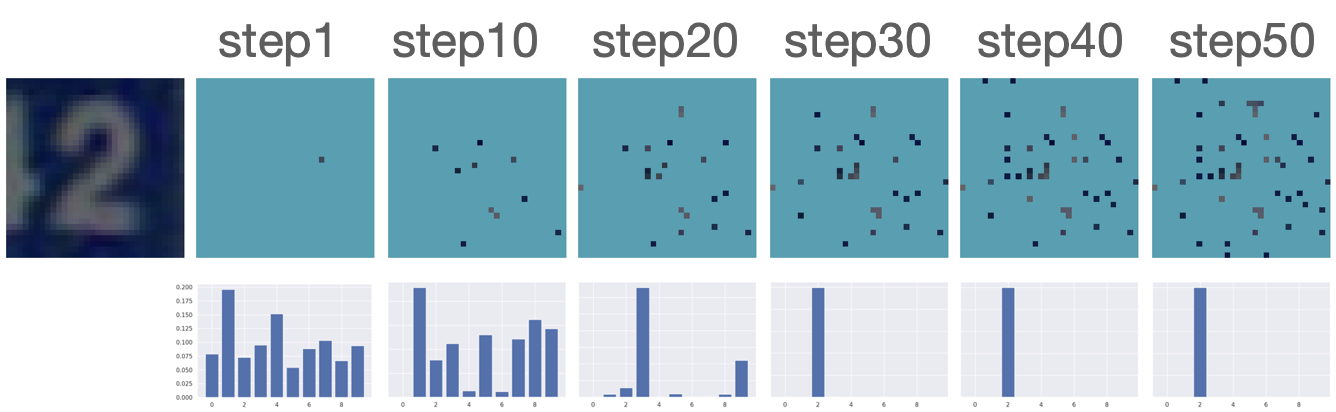}
    \vspace{-10pt}
    \caption{Examples of the acquisition process from our robust AFA framework. The bar charts demonstrate the prediction probability at the corresponding acquisition step.}
    \label{fig:rafa}
\end{wrapfigure}

\paragraph{Robust Active Feature Acquisition}
We first evaluate the AFA tasks using several classification datasets. The agent is trained to acquire the pixel values. For color images, the agent acquires all three channels at once. For MNIST \cite{lecun2010mnist} and FMNIST \cite{xiao2017/online}, we follow GSMRL to train the surrogate model using a class conditioned ACFlow \cite{li2020acflow}; for SVHN \cite{Netzer2011}, we simply use a partially observed classifier to learn $p(y \mid x_o)$ since we found ACFlow difficult to train for this dataset. The auxiliary information is accordingly modified to contain only the prediction probability. Figure~\ref{fig:acc} and \ref{fig:auc} report the classification accuracy and OOD detection AUROC respectively. The accuracy is significantly higher for RL approaches than the random acquisition policy. Although we expect a trade-off between accuracy and OOD detection performance for our robust AFA framework, the accuracy is actually comparable to GSMRL and sometimes even better across the datasets. Meanwhile, the OOD detection performance for our robust AFA framework is significantly improved by enforcing the agent to acquire informative features for OOD identification. For SVHN and CIFAR10 detection, the AUROC for GSMRL is even lower than the random policy, which we believe is because of the discrepancy of informative features for two different goals. Augmented with the detector reward solves the problem and improves the detection performance even further.
Figure~\ref{fig:rafa} presents several examples of the acquisition process from our robust AFA framework. We can see the prediction becomes certain after only a few acquisition steps. See appendix~\ref{sec:appendix_exp} for additional examples.

\begin{figure}
\begin{minipage}{0.49\linewidth}
    \centering
    \subfigure[MNIST]{\includegraphics[width=0.49\linewidth]{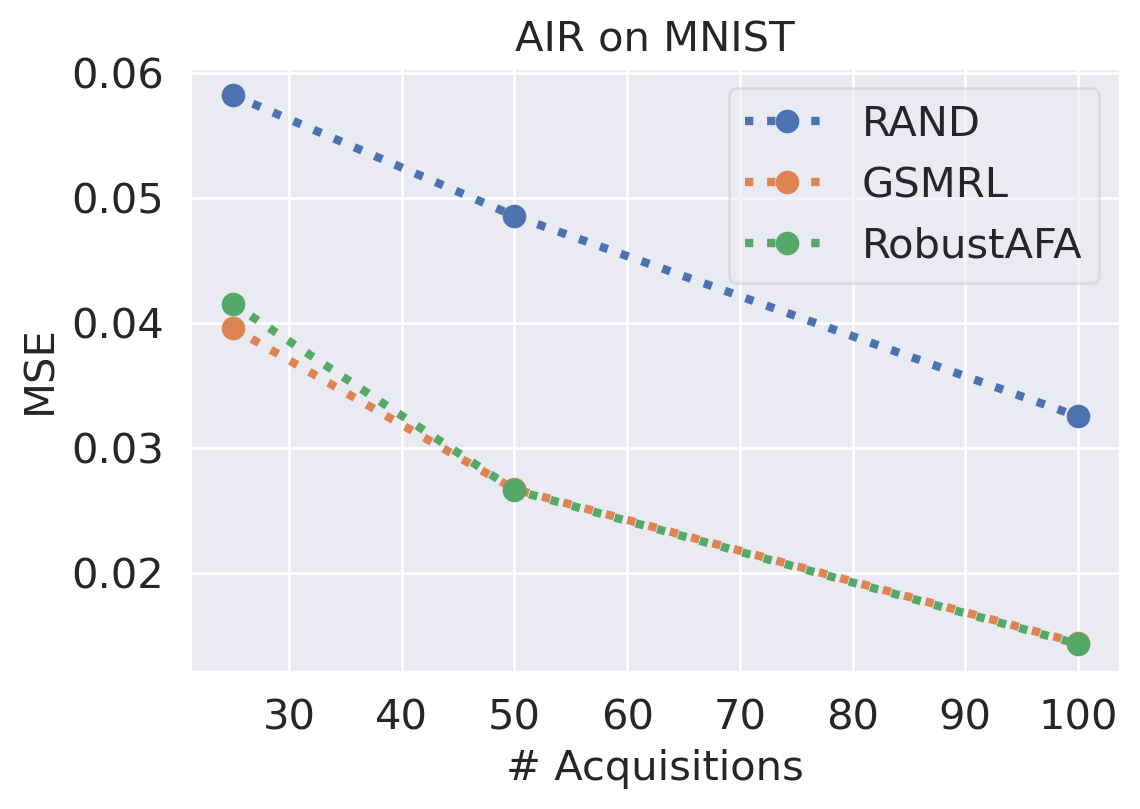}}
    \subfigure[FMNIST]{\includegraphics[width=0.49\linewidth]{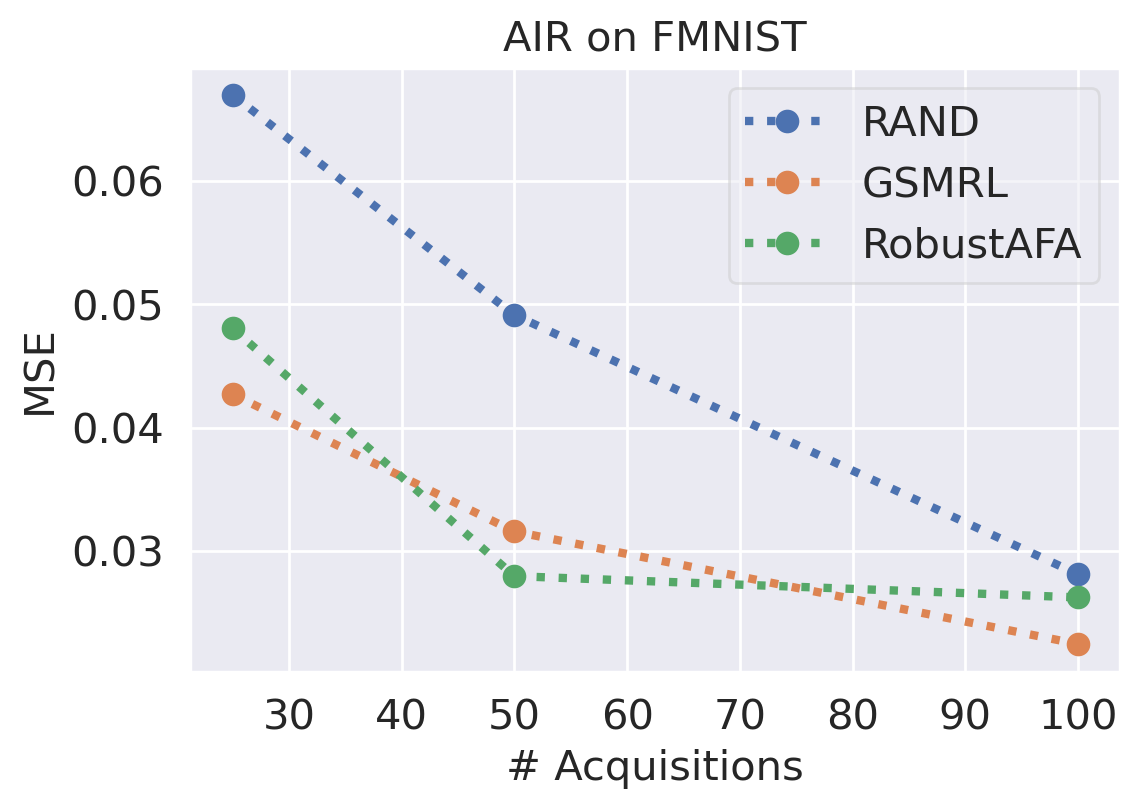}}
    \vspace{-5pt}
    \caption{Reconstruction MSE for robust AIR.}
    \label{fig:mse}
\end{minipage}
\begin{minipage}{0.49\linewidth}
    \centering
    \subfigure[MNIST - Omniglot]{\includegraphics[width=0.49\linewidth]{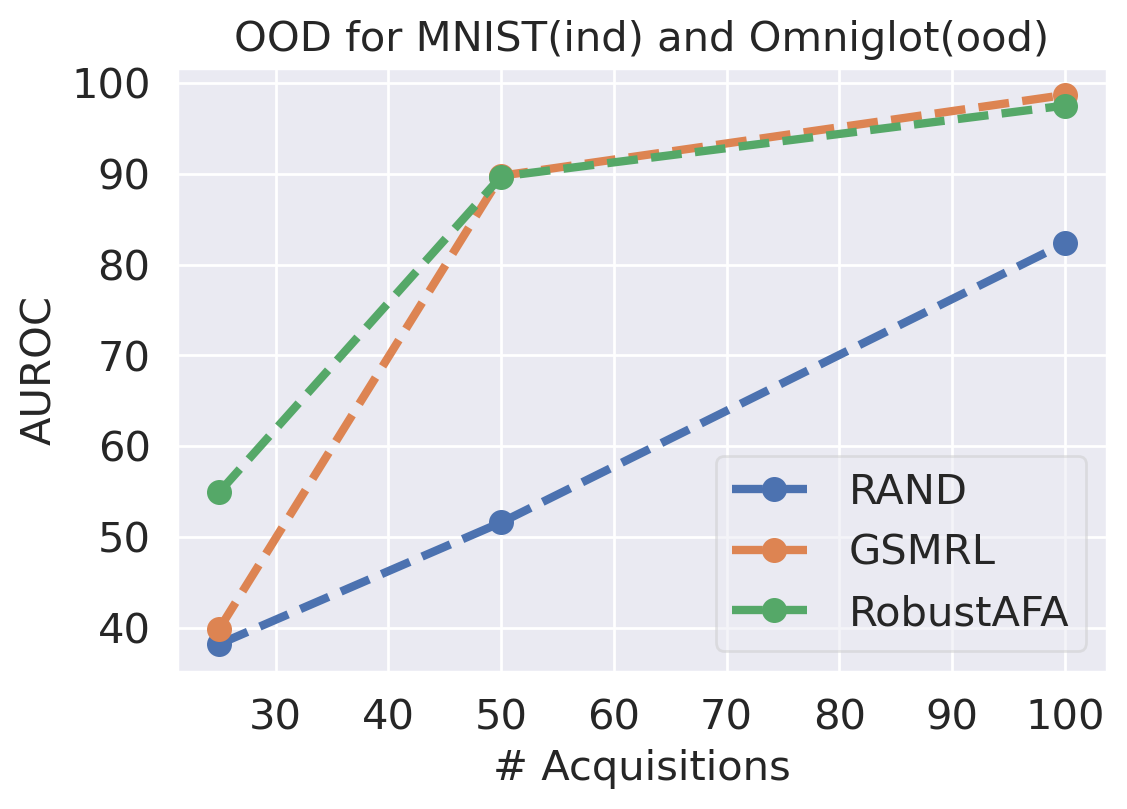}}
    \subfigure[FMNIST - MNIST]{\includegraphics[width=0.49\linewidth]{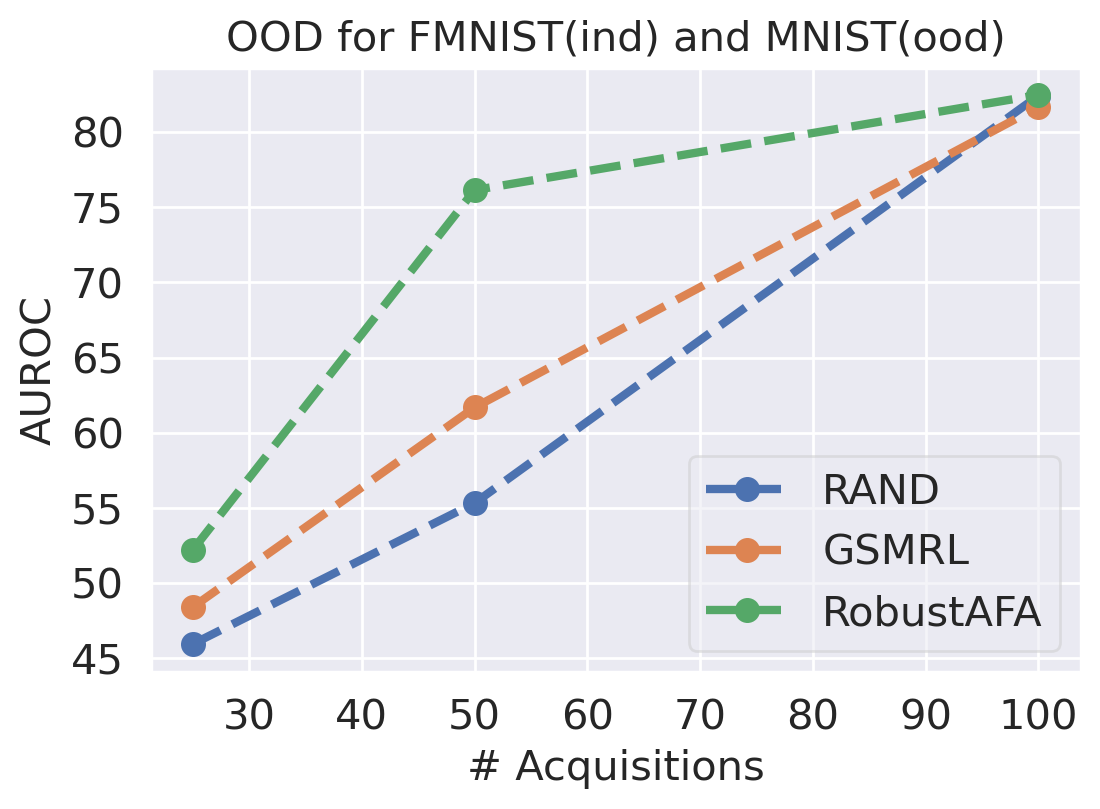}}
    \vspace{-7pt}
    \caption{OOD detection for robust AIR.}
    \label{fig:rec}
\end{minipage}
\vspace{-10pt}
\end{figure}

\begin{figure}
    \centering
    \includegraphics[width=0.49\linewidth]{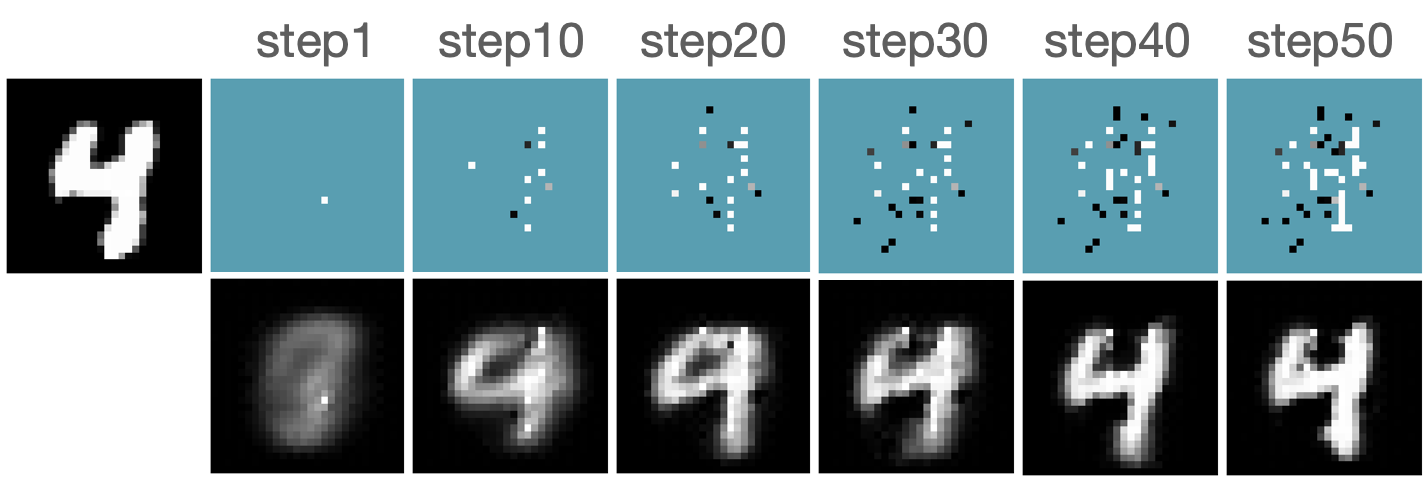}
    \includegraphics[width=0.49\linewidth]{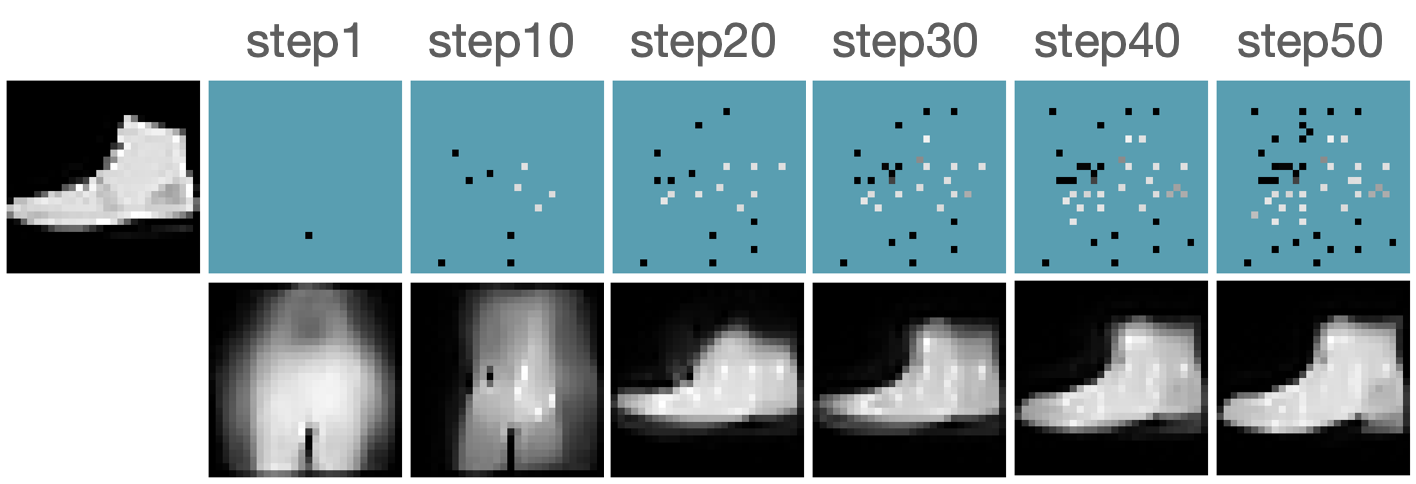}
    \vspace{-5pt}
    \caption{Examples of the acquisition process for robust AIR.}
    \label{fig:rair}
    \vspace{-8pt}
\end{figure}

\paragraph{Robust Active Instance Recognition}
In this section, we evaluate the AIR task using MNIST and FashionMNSIT datasets. Following GSMRL \cite{li2020active}, we use ACFlow as the surrogate model. Figure~\ref{fig:mse} and \ref{fig:rec} report the reconstruction MSE and OOD detection performance respectively using the acquired features. We can see our robust AIR framework improves the OOD detection performance significantly, especially when the acquisition budget is low, while the reconstruction MSEs are comparable to GSMRL. Figure~\ref{fig:rair} presents several examples of the acquisition process for robust AIR.

\begin{wrapfigure}{r}{0.3\linewidth}
    \centering
    \vspace{-10pt}
    \includegraphics[width=\linewidth]{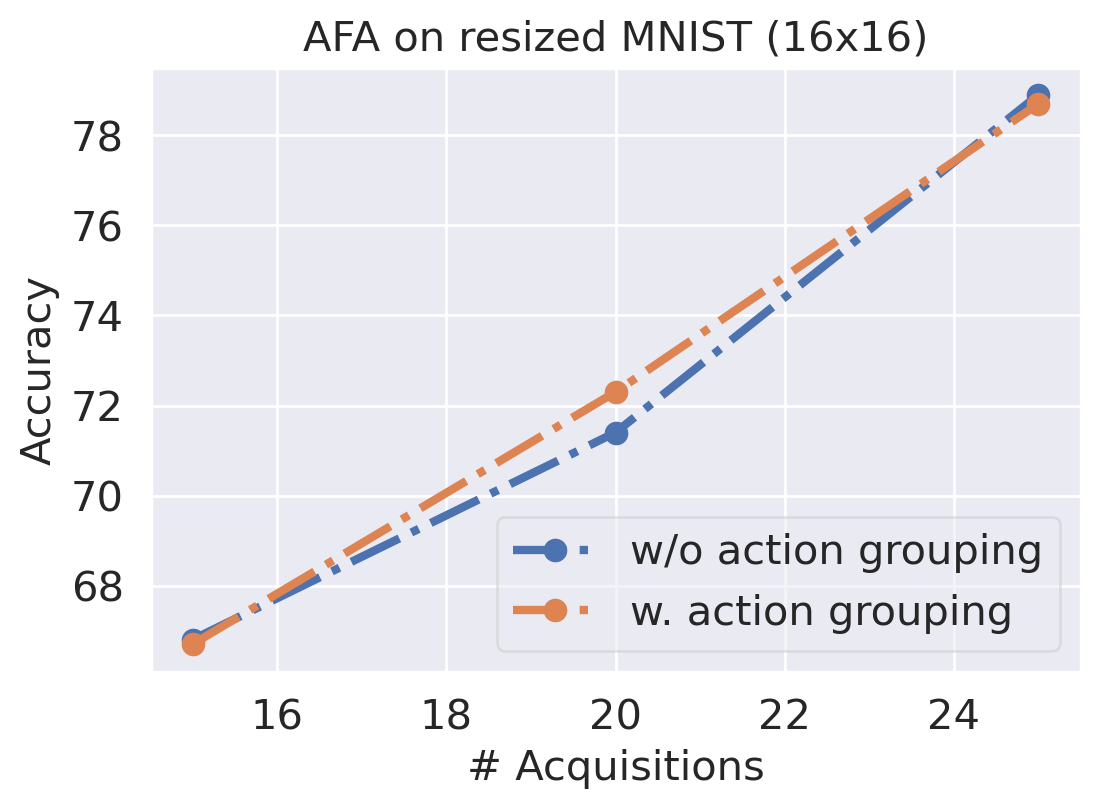}
    \vspace{-13pt}
    \caption{Compare AFA performance with or without action grouping.}
    \label{fig:ablation_acc}
\end{wrapfigure}

\paragraph{Ablations}
Our proposed action grouping technique enables the agent to acquire features from a potentially large pool of candidates. However, it also introduces some complexity due to the autoregressive factorization in \eqref{eq:autoreg_action}. In Fig.~\ref{fig:ablation_acc}, we compare two agents with and without the action grouping using a downsampled MNIST. We can see the action grouping does not degrade the performance on smaller dimensionalities whilst allowing one to work over larger dimensionalities that previous methods cannot scale to.

Although our PO-MSMA is designed for partially observed instances, it can handle fully observed ones as special cases. In Table~\ref{tab:ablation_pomsma}, we report the AUROC scores for both methods. We can see our PO-MSMA is competitive even though it is not trained to detect fully observed instances.

\begin{table}[H]
    \centering
    \vspace{-8pt}
    \caption{Comparison with MSMA for fully observed OOD detection. AUROC scores are reported.}
    \label{tab:ablation_pomsma}
    \small
    \begin{tabular}{c|c|c|c|c}
    \toprule
         &  MNIST - Omniglot & FMNIST - MNIST & SVHN - CIFAR10 & CIFAR10 - SVHN\\
    \midrule
        MSMA & - & 82.56 & 97.60 & 95.50 \\
        PO-MSMA & 99.55 & 96.62 & 97.77 & 74.74 \\
    \bottomrule
    \end{tabular}
\end{table}

\section{Discussion and Conclusion}\label{sec:conclusion}
In this work, we investigate an understudied problem in AFA, where we  switch gears from acquiring informative features to acquiring robust features. A robust AFA framework is proposed here to acquire feature actively and determine whether the input is OOD using only the acquired subset of features. In order to scale up the AFA models to practical use, we develop a hierarchical acquisition policy, where the candidate features are grouped together based on their relevance to the target. Our framework represents the first AFA model that can deal with a potentially large pool of candidate features. Extensive experiments are conducted to showcase the effectiveness of our framework. Due to the fact that instances are partially observed for AFA, our framework is not guaranteed to be robust for adversaries (a particular type of OOD with minimum modification to a valid input), since the adversary can easily modify those unobserved features to fool the detector. However, the existence of an adversarially robust AFA model is also an open question.


\bibliographystyle{unsrt}
\bibliography{neurips_2021}

\begin{thebibliography}{10}

\bibitem{li2020active}
Yang Li and Junier~B Oliva.
\newblock Active feature acquisition with generative surrogate models.
\newblock {\em arXiv preprint arXiv:2010.02433}, 2020.

\bibitem{shim2018joint}
Hajin Shim, Sung~Ju Hwang, and Eunho Yang.
\newblock Joint active feature acquisition and classification with
  variable-size set encoding.
\newblock {\em Advances in neural information processing systems},
  31:1368--1378, 2018.

\bibitem{dulac2015deep}
Gabriel Dulac-Arnold, Richard Evans, Hado van Hasselt, Peter Sunehag, Timothy
  Lillicrap, Jonathan Hunt, Timothy Mann, Theophane Weber, Thomas Degris, and
  Ben Coppin.
\newblock Deep reinforcement learning in large discrete action spaces.
\newblock {\em arXiv preprint arXiv:1512.07679}, 2015.

\bibitem{ling2004decision}
Charles~X Ling, Qiang Yang, Jianning Wang, and Shichao Zhang.
\newblock Decision trees with minimal costs.
\newblock In {\em Proceedings of the twenty-first international conference on
  Machine learning}, page~69, 2004.

\bibitem{chai2004test}
Xiaoyong Chai, Lin Deng, Qiang Yang, and Charles~X Ling.
\newblock Test-cost sensitive naive bayes classification.
\newblock In {\em Fourth IEEE International Conference on Data Mining
  (ICDM'04)}, pages 51--58. IEEE, 2004.

\bibitem{nan2014fast}
Feng Nan, Joseph Wang, Kirill Trapeznikov, and Venkatesh Saligrama.
\newblock Fast margin-based cost-sensitive classification.
\newblock In {\em 2014 IEEE International Conference on Acoustics, Speech and
  Signal Processing (ICASSP)}, pages 2952--2956. IEEE, 2014.

\bibitem{ma2018eddi}
Chao Ma, Sebastian Tschiatschek, Konstantina Palla, Jos{\'e}~Miguel
  Hern{\'a}ndez-Lobato, Sebastian Nowozin, and Cheng Zhang.
\newblock Eddi: Efficient dynamic discovery of high-value information with
  partial vae.
\newblock {\em arXiv preprint arXiv:1809.11142}, 2018.

\bibitem{bernardo1979expected}
Jos{\'e}~M Bernardo.
\newblock Expected information as expected utility.
\newblock {\em the Annals of Statistics}, pages 686--690, 1979.

\bibitem{gong2019icebreaker}
Wenbo Gong, Sebastian Tschiatschek, Sebastian Nowozin, Richard~E Turner,
  Jos{\'e}~Miguel Hern{\'a}ndez-Lobato, and Cheng Zhang.
\newblock Icebreaker: Element-wise efficient information acquisition with a
  bayesian deep latent gaussian model.
\newblock 2019.

\bibitem{zubek2004pruning}
Valentina~Bayer Zubek, Thomas~Glen Dietterich, et~al.
\newblock Pruning improves heuristic search for cost-sensitive learning.
\newblock 2004.

\bibitem{ruckstiess2011sequential}
Thomas R{\"u}ckstie{\ss}, Christian Osendorfer, and Patrick van~der Smagt.
\newblock Sequential feature selection for classification.
\newblock In {\em Australasian joint conference on artificial intelligence},
  pages 132--141. Springer, 2011.

\bibitem{he2012imitation}
He~He, Jason Eisner, and Hal Daume.
\newblock Imitation learning by coaching.
\newblock {\em Advances in Neural Information Processing Systems},
  25:3149--3157, 2012.

\bibitem{he2016active}
He~He, Paul Mineiro, and Nikos Karampatziakis.
\newblock Active information acquisition.
\newblock {\em arXiv preprint arXiv:1602.02181}, 2016.

\bibitem{minsky1961steps}
Marvin Minsky.
\newblock Steps toward artificial intelligence.
\newblock {\em Proceedings of the IRE}, 49(1):8--30, 1961.

\bibitem{sutton1988learning}
Richard~S Sutton.
\newblock Learning to predict by the methods of temporal differences.
\newblock {\em Machine learning}, 3(1):9--44, 1988.

\bibitem{li2020acflow}
Yang Li, Shoaib Akbar, and Junier Oliva.
\newblock {ACF}low: Flow models for arbitrary conditional likelihoods.
\newblock In {\em International Conference on Machine Learning}, pages
  5831--5841. PMLR, 2020.

\bibitem{pineda2020active}
Luis Pineda, Sumana Basu, Adriana Romero, Roberto Calandra, and Michal
  Drozdzal.
\newblock Active mr k-space sampling with reinforcement learning.
\newblock In {\em International Conference on Medical Image Computing and
  Computer-Assisted Intervention}, pages 23--33. Springer, 2020.

\bibitem{bakker2020experimental}
Tim Bakker, Herke van Hoof, and Max Welling.
\newblock Experimental design for mri by greedy policy search.
\newblock {\em Advances in Neural Information Processing Systems}, 33, 2020.

\bibitem{zhang2019reducing}
Zizhao Zhang, Adriana Romero, Matthew~J Muckley, Pascal Vincent, Lin Yang, and
  Michal Drozdzal.
\newblock Reducing uncertainty in undersampled mri reconstruction with active
  acquisition.
\newblock In {\em Proceedings of the IEEE/CVF Conference on Computer Vision and
  Pattern Recognition}, pages 2049--2058, 2019.

\bibitem{gorp2021active}
Hans van Gorp, Iris~A.M. Huijben, Bastiaan~S. Veeling, Nicola Pezzotti, and
  Ruud~Van Sloun.
\newblock Active deep probabilistic subsampling, 2021.

\bibitem{ovadia2019can}
Yaniv Ovadia, Emily Fertig, Jie Ren, Zachary Nado, David Sculley, Sebastian
  Nowozin, Joshua~V Dillon, Balaji Lakshminarayanan, and Jasper Snoek.
\newblock Can you trust your model's uncertainty? evaluating predictive
  uncertainty under dataset shift.
\newblock {\em arXiv preprint arXiv:1906.02530}, 2019.

\bibitem{kumar2019verified}
Ananya Kumar, Percy Liang, and Tengyu Ma.
\newblock Verified uncertainty calibration.
\newblock {\em arXiv preprint arXiv:1909.10155}, 2019.

\bibitem{blundell2015weight}
Charles Blundell, Julien Cornebise, Koray Kavukcuoglu, and Daan Wierstra.
\newblock Weight uncertainty in neural network.
\newblock In {\em International Conference on Machine Learning}, pages
  1613--1622. PMLR, 2015.

\bibitem{lakshminarayanan2016simple}
Balaji Lakshminarayanan, Alexander Pritzel, and Charles Blundell.
\newblock Simple and scalable predictive uncertainty estimation using deep
  ensembles.
\newblock {\em arXiv preprint arXiv:1612.01474}, 2016.

\bibitem{gal2016dropout}
Yarin Gal and Zoubin Ghahramani.
\newblock Dropout as a bayesian approximation: Representing model uncertainty
  in deep learning.
\newblock In {\em international conference on machine learning}, pages
  1050--1059. PMLR, 2016.

\bibitem{van2020uncertainty}
Joost Van~Amersfoort, Lewis Smith, Yee~Whye Teh, and Yarin Gal.
\newblock Uncertainty estimation using a single deep deterministic neural
  network.
\newblock In {\em International Conference on Machine Learning}, pages
  9690--9700. PMLR, 2020.

\bibitem{lecun1998gradient}
Yann LeCun, L{\'e}on Bottou, Yoshua Bengio, and Patrick Haffner.
\newblock Gradient-based learning applied to document recognition.
\newblock {\em Proceedings of the IEEE}, 86(11):2278--2324, 1998.

\bibitem{liu2020simple}
Jeremiah~Zhe Liu, Zi~Lin, Shreyas Padhy, Dustin Tran, Tania Bedrax-Weiss, and
  Balaji Lakshminarayanan.
\newblock Simple and principled uncertainty estimation with deterministic deep
  learning via distance awareness.
\newblock {\em arXiv preprint arXiv:2006.10108}, 2020.

\bibitem{van2021improving}
Joost van Amersfoort, Lewis Smith, Andrew Jesson, Oscar Key, and Yarin Gal.
\newblock Improving deterministic uncertainty estimation in deep learning for
  classification and regression.
\newblock {\em arXiv preprint arXiv:2102.11409}, 2021.

\bibitem{bishop1994novelty}
Christopher~M Bishop.
\newblock Novelty detection and neural network validation.
\newblock {\em IEE Proceedings-Vision, Image and Signal processing},
  141(4):217--222, 1994.

\bibitem{nalisnick2018do}
Eric Nalisnick, Akihiro Matsukawa, Yee~Whye Teh, Dilan Gorur, and Balaji
  Lakshminarayanan.
\newblock Do deep generative models know what they don't know?
\newblock In {\em International Conference on Learning Representations}, 2019.

\bibitem{hendrycks2018deep}
Dan Hendrycks, Mantas Mazeika, and Thomas Dietterich.
\newblock Deep anomaly detection with outlier exposure.
\newblock In {\em International Conference on Learning Representations}, 2019.

\bibitem{choi2018waic}
Hyunsun Choi, Eric Jang, and Alexander~A Alemi.
\newblock Waic, but why? generative ensembles for robust anomaly detection.
\newblock {\em arXiv preprint arXiv:1810.01392}, 2018.

\bibitem{ren2019likelihood}
Jie Ren, Peter~J Liu, Emily Fertig, Jasper Snoek, Ryan Poplin, Mark~A DePristo,
  Joshua~V Dillon, and Balaji Lakshminarayanan.
\newblock Likelihood ratios for out-of-distribution detection.
\newblock {\em arXiv preprint arXiv:1906.02845}, 2019.

\bibitem{nalisnick2019detecting}
Eric Nalisnick, Akihiro Matsukawa, Yee~Whye Teh, and Balaji Lakshminarayanan.
\newblock Detecting out-of-distribution inputs to deep generative models using
  typicality.
\newblock {\em arXiv preprint arXiv:1906.02994}, 2019.

\bibitem{morningstar2021density}
Warren Morningstar, Cusuh Ham, Andrew Gallagher, Balaji Lakshminarayanan, Alex
  Alemi, and Joshua Dillon.
\newblock Density of states estimation for out of distribution detection.
\newblock In {\em International Conference on Artificial Intelligence and
  Statistics}, pages 3232--3240. PMLR, 2021.

\bibitem{mahmood2021multiscale}
Ahsan Mahmood, Junier Oliva, and Martin~Andreas Styner.
\newblock Multiscale score matching for out-of-distribution detection.
\newblock In {\em International Conference on Learning Representations}, 2021.

\bibitem{song2019generative}
Yang Song and Stefano Ermon.
\newblock Generative modeling by estimating gradients of the data distribution.
\newblock {\em arXiv preprint arXiv:1907.05600}, 2019.

\bibitem{majeed2020exact}
Sultan~Javed Majeed and Marcus Hutter.
\newblock Exact reduction of huge action spaces in general reinforcement
  learning.
\newblock {\em arXiv preprint arXiv:2012.10200}, 2020.

\bibitem{papamakarios2017masked}
George Papamakarios, Theo Pavlakou, and Iain Murray.
\newblock Masked autoregressive flow for density estimation.
\newblock {\em arXiv preprint arXiv:1705.07057}, 2017.

\bibitem{lecun2010mnist}
Yann LeCun, Corinna Cortes, and CJ~Burges.
\newblock Mnist handwritten digit database.
\newblock {\em ATT Labs [Online]. Available: http://yann.lecun.com/exdb/mnist},
  2, 2010.

\bibitem{xiao2017/online}
Han Xiao, Kashif Rasul, and Roland Vollgraf.
\newblock Fashion-mnist: a novel image dataset for benchmarking machine
  learning algorithms, 2017.

\bibitem{Netzer2011}
Yuval Netzer, Tao Wang, Adam Coates, Alessandro Bissacco, Bo~Wu, and Andrew~Y
  Ng.
\newblock Reading digits in natural images with unsupervised feature learning.
\newblock 2011.

\end{thebibliography}

\section*{Checklist}

\begin{enumerate}

\item For all authors...
\begin{enumerate}
  \item Do the main claims made in the abstract and introduction accurately reflect the paper's contributions and scope?
    \answerYes{}
  \item Did you describe the limitations of your work?
    \answerYes{See Sec.~\ref{sec:conclusion}}
  \item Did you discuss any potential negative societal impacts of your work?
    \answerYes{See appendix \ref{sec:impact}}
  \item Have you read the ethics review guidelines and ensured that your paper conforms to them?
    \answerYes{}
\end{enumerate}

\item If you are including theoretical results...
\begin{enumerate}
  \item Did you state the full set of assumptions of all theoretical results?
    \answerNA{}
	\item Did you include complete proofs of all theoretical results?
    \answerNA{}
\end{enumerate}

\item If you ran experiments...
\begin{enumerate}
  \item Did you include the code, data, and instructions needed to reproduce the main experimental results (either in the supplemental material or as a URL)?
    \answerYes{Code is provided as supplemental material.}
  \item Did you specify all the training details (e.g., data splits, hyperparameters, how they were chosen)?
    \answerYes{See appendix \ref{sec:appendix_exp}}
	\item Did you report error bars (e.g., with respect to the random seed after running experiments multiple times)?
    \answerYes{See appendix \ref{sec:appendix_exp}}
	\item Did you include the total amount of compute and the type of resources used (e.g., type of GPUs, internal cluster, or cloud provider)?
    \answerYes{See appendix \ref{sec:appendix_exp}}
\end{enumerate}

\item If you are using existing assets (e.g., code, data, models) or curating/releasing new assets...
\begin{enumerate}
  \item If your work uses existing assets, did you cite the creators?
    \answerYes{All datasets used in this work are publically available.}
  \item Did you mention the license of the assets?
    \answerNA{}
  \item Did you include any new assets either in the supplemental material or as a URL?
    \answerNA{}
  \item Did you discuss whether and how consent was obtained from people whose data you're using/curating?
    \answerNA{}
  \item Did you discuss whether the data you are using/curating contains personally identifiable information or offensive content?
    \answerNA{}
\end{enumerate}

\item If you used crowdsourcing or conducted research with human subjects...
\begin{enumerate}
  \item Did you include the full text of instructions given to participants and screenshots, if applicable?
    \answerNA{}
  \item Did you describe any potential participant risks, with links to Institutional Review Board (IRB) approvals, if applicable?
    \answerNA{}
  \item Did you include the estimated hourly wage paid to participants and the total amount spent on participant compensation?
    \answerNA{}
\end{enumerate}

\end{enumerate}


\clearpage
\appendix

\renewcommand\thefigure{\thesection.\arabic{figure}}
\renewcommand\thetable{\thesection.\arabic{table}}
\renewcommand{\theequation}{\thesection.\arabic{equation}}

\setcounter{figure}{0} 
\setcounter{subfigure}{0}
\setcounter{equation}{0}

\section{Broader Impact}\label{sec:impact}
Our proposed robust AFA framework improves the existing AFA approaches by making it more robust to out-of-distribution inputs. It inherits the potentially broad application of AFA. For instance, it could inform autonomous vehicles when it is possible to accurately make an early prediction based on the limited data collected. In these time-critical applications, AFA have the potential to save lives. AFA will make a direct impact on applications where the computer is actively interacting with an user to gather information and make an assessment. Examples include surveys and customer service chat-bots, where AFA agent would determine the most informative next question to quickly and accurately make predictions based on the limited answers collected. Related applications of interest are AI education systems that personalize curricula and materials to each student. AFA agent could quickly assess a student’s current knowledge and abilities with a limited number of questions to expedite and personalize the curriculum. AFA will also have a deep impact for applications that collect data out in the field. For instance, in environmental applications agents must physically collect samples from various locations to assess if a region has been contaminated. AFA could reduce the number of locations sampled, and suggest the most informative locations to query with autonomous agents. AFA approaches also have the potential to protect privacy, since those private features can be removed from the candidate set beforehand and will never be acquired by the agent.

In addition to the positive impact, our robust AFA framework also have the same negative impact as AFA approaches. Due to the fact that features are partially observed during the acquisition process, it can be vulnerable to adversarial attacks, since an adversary can easily modify the unobserved features to fool the AFA agent.

\section{Experimental Details}\label{sec:appendix_exp}
\subsection{Robust Active Feature Acquisition}
\paragraph{Datasets} We evaluate the performance for robust AFA using several classification datasets. MNIST and FashionMNIST are two gray-scale image datasets of size $28\times28$, and SVHN is a color image dataset of size $32\times32$. Our framework acquires one pixel value at each acquisition step. For color images, it acquires all three channels at once.

\paragraph{Dynamics Model} For MNIST and FashionMNIST, we follow GSMRL \cite{li2020active} to use a class conditioned ACFlow for dynamics modeling. ACFlow captures the arbitrary conditional distribution $p(x_u \mid x_o, y)$, and the prediction using an arbitrary subset can be derived using the Bayes rule, i.e.,
\begin{equation}
    p(y \mid x_o) = \frac{p(x_o \mid y)p(y)}{\sum_{y'}p(x_o \mid y')p(y')}.
\end{equation}
The architecture of ACFlow closely follows GSMRL, which contains a stack of affine coupling layers and a Gaussian base likelihood module. The dynamics model is used to assess the intermediate reward of an acquisition and provide auxiliary information to assist the agent. please refer to Sec.~\ref{sec:background} for details.

For SVHN, it is hard for ACFlow to balance the likelihood objective and the classification loss. Instead, we use a simple classifier with partially observed inputs to learn $p(y \mid x_o)$. We use the ResNet50 architecture and modify it to take in masked inputs and a binary mask. During training, we sample the mask at random. Since the partially observed classifier does not explicitly capture the dependencies among features, it cannot provide any prediction about the unobserved features. Although, it can still assess the intermediate reward using the information gain. We also use the current prediction probability as the auxiliary information.

\paragraph{PO-MSMA} Our PO-MSMA model consists of a score matching network and a likelihood model over the norm of scores. We modify the original NCSN model to produce the scores for arbitrary marginal distributions $\nabla_{\tilde{x}_m} \log p(\tilde{x}_m)$. Specifically, the inputs contain the masked images $x_o$ and a binary mask indicating the observed pixels. The output is a tensor with the same size as the input image. We then mask out the unobserved dimensions for the output and compute the norm only for those observed pixels. Throughout the experiment, we use 10 noise scales, thus obtaining 10 summary statistics, $s_1, \ldots, s_{10}$, for each input image. Then, we train a conditional autoregressive model for the norms conditioned on the binary mask $\I_m$, i.e., 
\begin{equation}
    p(s_1, \ldots,s_{10} \mid \I_m).
\end{equation}
Given a test image $x$ with observed dimensions $m$, the OOD detection starts by calculating the norm of scores on different noise levels. Then, the conditional likelihood $p(s_1,\ldots,s_{10} \mid \I_m)$ is thresholded to determine whether the given partially observed input is OOD or not. Here, we report the AUROC scores to evaluate the OOD detection performance. 

\paragraph{AFA Agent} We use PPO algorithm to train our AFA agent. Given the observed dimensions as the state, we first use a two-layer convolutional network with max pooling to extract an embedding, from which the actor and critic are derived using two fully connected layers. The actor network predicts the probability of the next action, where the probabilities of observed features are manually set to zero. The critic network is used to estimate the state values. The AFA agent observes the current acquired features and determines which next feature to acquire. It stops acquiring more features when the acquired features exceed the acquisition budget. Throughout this work, we assume each feature has the same cost, thus the acquisition budget is equivalent to the number of features to be acquired. In GSMRL \cite{li2020active}, the authors also learn a predictor along with the agent. However, we did not find it beneficial at the early stage of experiment. Therefore, we directly use the dynamics model to make a final prediction.

\paragraph{Baselines} Both greedy and RL based approaches have been proposed for the AFA task. However, they all deal with a small number of candidate features and have difficulty scaling to large ones. For example, the greedy approach, EDDI \cite{ma2018eddi}, has a $O(Nd)$ computation complexity for acquiring $N$ features from a $d$ dimensional feature space. Model-free and model-based RL approaches are known difficult dealing with large action spaces \cite{li2020active}. In order to evaluate the OOD detection performance using commonly used benchmarks, we modify the state-of-the-art AFA approach, GSMRL \cite{li2020active}, with our proposed action space grouping technique. We also compare to a random policy where a random unobserved feature is acquired at each acquisition step.

\begin{figure}
\begin{minipage}{0.49\linewidth}
    \centering
    \includegraphics[width=\linewidth]{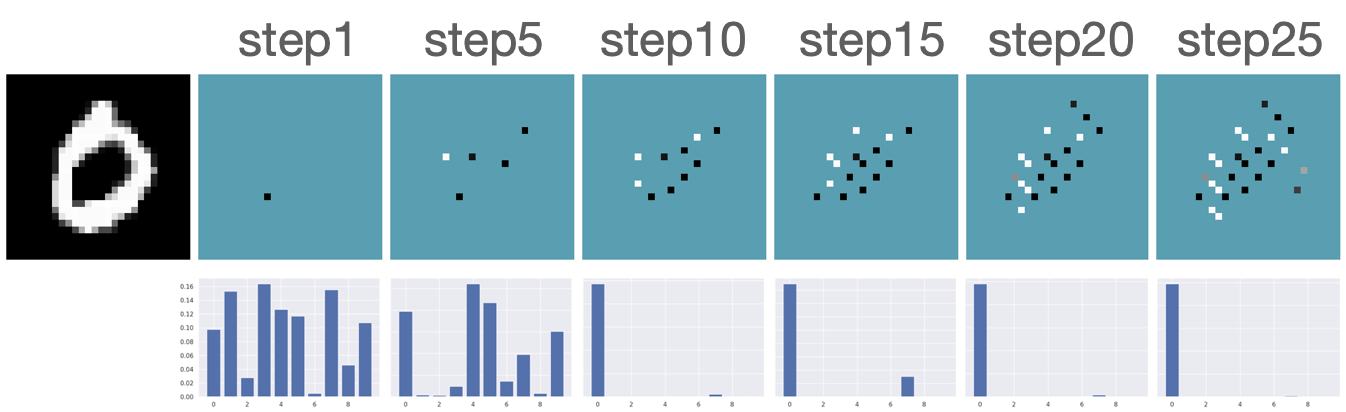}
    
    \vspace{5pt}
    
    \includegraphics[width=\linewidth]{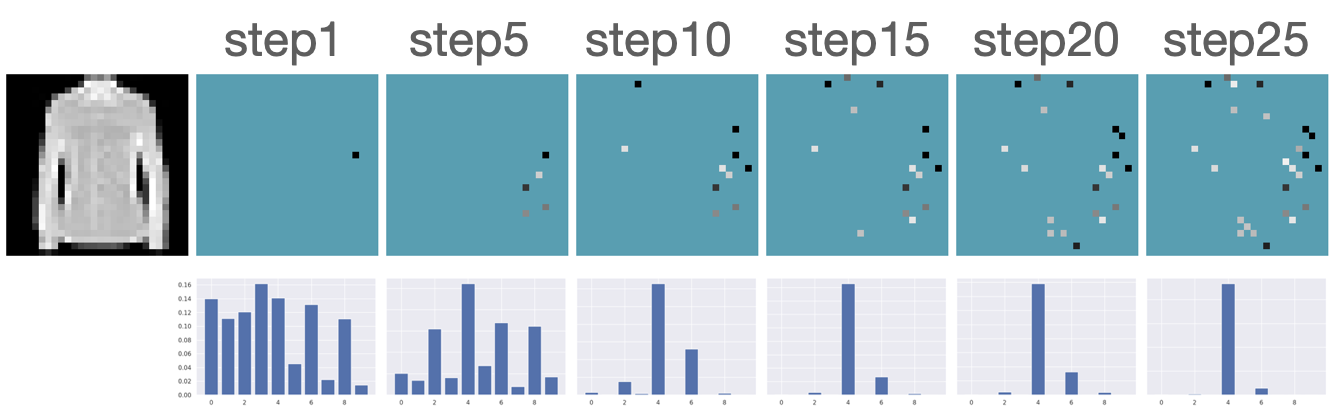}
    
    \vspace{5pt}
    
    \includegraphics[width=\linewidth]{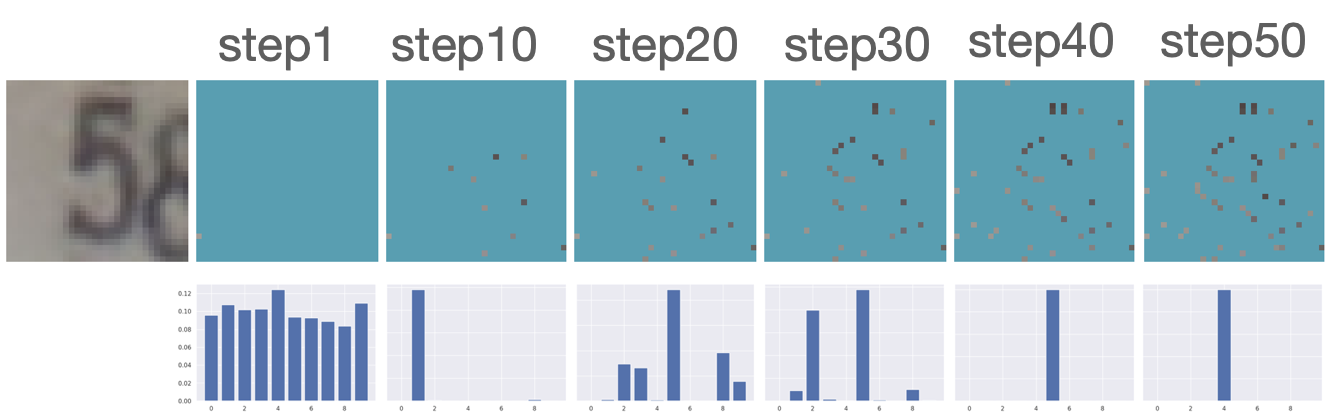}
    \caption{Additional results from our robust AFA framework.}
    \label{fig:appendix_rafa}
\end{minipage}
\hfill
\begin{minipage}{0.46\linewidth}
    \centering
    \includegraphics[width=\linewidth]{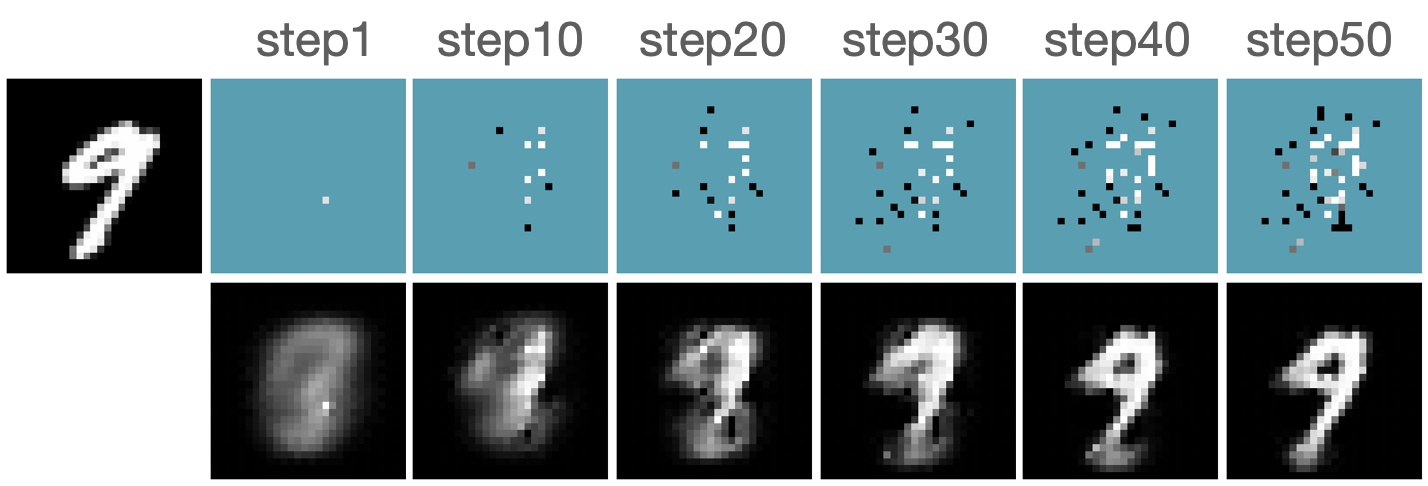}
    
    \vspace{5pt}
    
    \includegraphics[width=\linewidth]{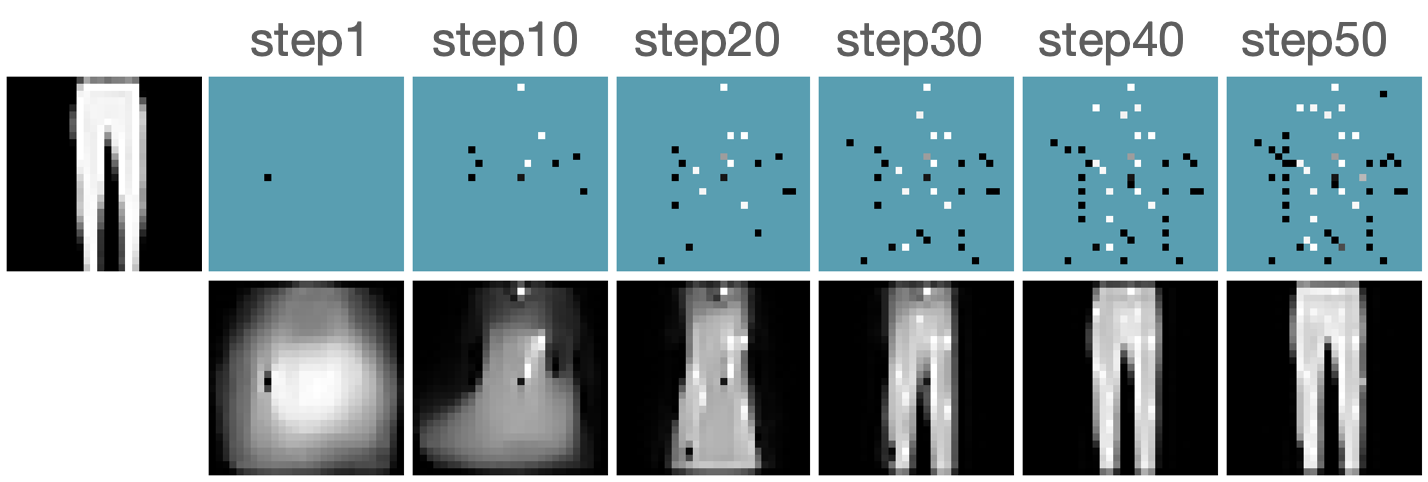}
    
    \vspace{5pt}
    
    \includegraphics[width=\linewidth]{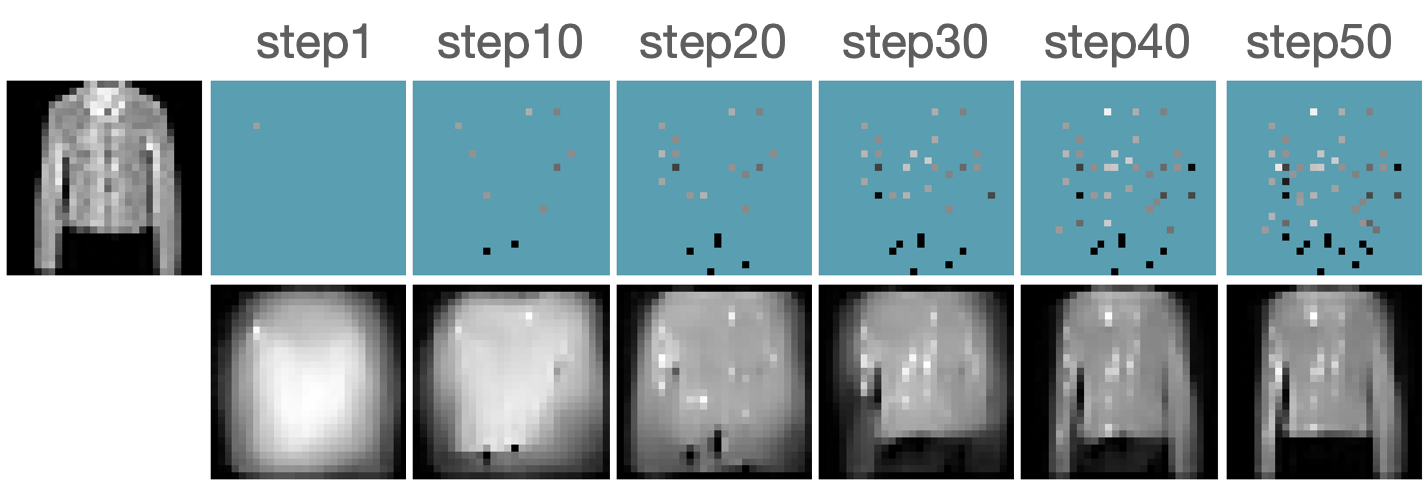}
    \caption{Additional results from our robust AIR framework.}
    \label{fig:appendix_rair}
\end{minipage}
\end{figure}

\paragraph{Additional Results} Figure \ref{fig:appendix_rafa} presents additional results for acquiring features using our robust AFA framework. Our model successfully recognizes the underlying classes using only a small subset of features.

\subsection{Robust Active Instance Recognition}
\paragraph{Datasets} We evaluate the AIR performance using MNIST and FashionMNIST. The model acquires one pixel at each acquisition step to reconstruct those unobserved pixels.

\paragraph{Dynamics Model} Following GSMRL \cite{li2020active}, we use ACFlow to model the dynamics. Specifically, ACFlow learns the arbitrary conditional distribution $p(x_u \mid x_o)$, and the prediction about the unobserved pixels are simply sampled from this distribution. The intermediate reward is defined as the improvement of the log likelihood per dimension, i.e.,
\begin{equation}
    r_m(s, i) = \frac{\log p(x_{u \setminus i} \mid x_o)}{|u|-1} - \frac{\log p(x_u \mid x_o)}{|u|}.
\end{equation}
The dynamics model also provides auxiliary information to the agent, which contains the predicted mean and variance of the unobserved pixels.

\paragraph{PO-MSMA} The OOD detector is the same as used in the AFA task.

\paragraph{AIR Agent} The agent is also the same as the AFA task, except the final reward is given as the MSE between the prediction and the groundtruth.

\paragraph{Baselines} Similar to the AFA task described above, we compare to a modified GSMRL algorithm and a random acquisition policy.

\paragraph{Additional Results} Figure \ref{fig:appendix_rair} presents several acquisition processes for the AIR task using our proposed framework. The prediction quickly becomes certain after only several acquisitions.

\subsection{Ablations}
\begin{figure}
    \centering
    \includegraphics[width=0.4\linewidth]{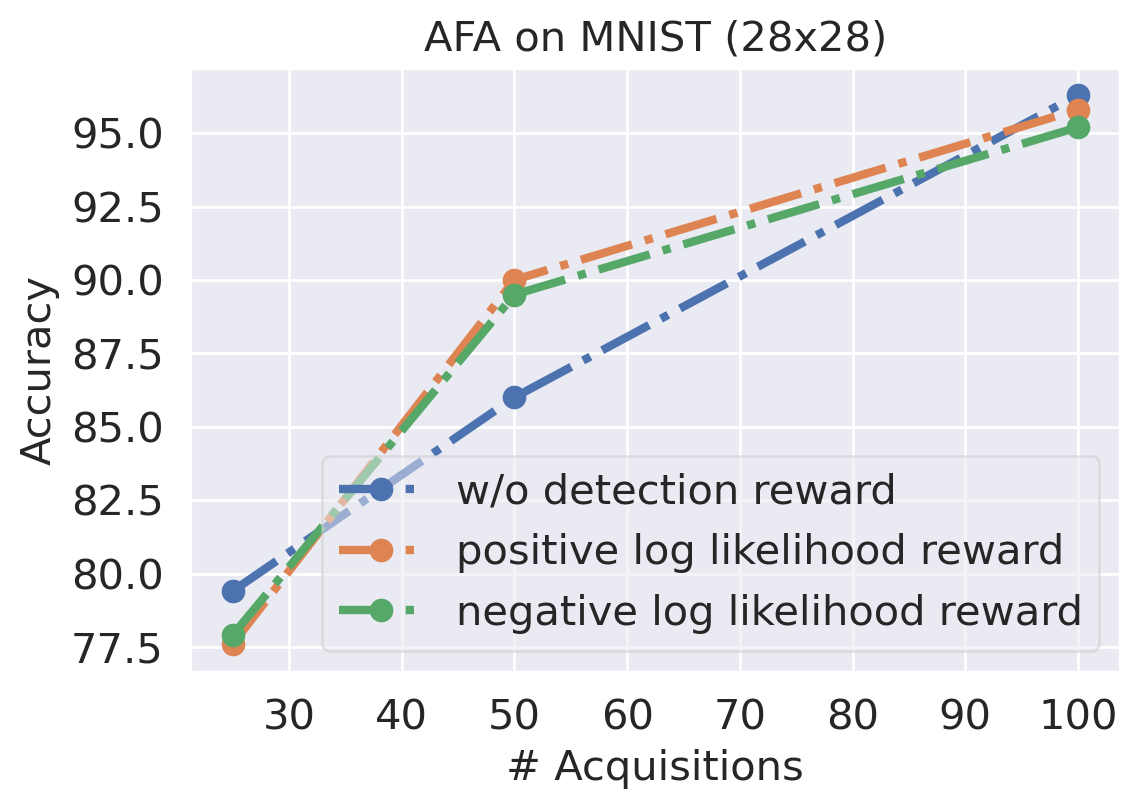}
    \includegraphics[width=0.4\linewidth]{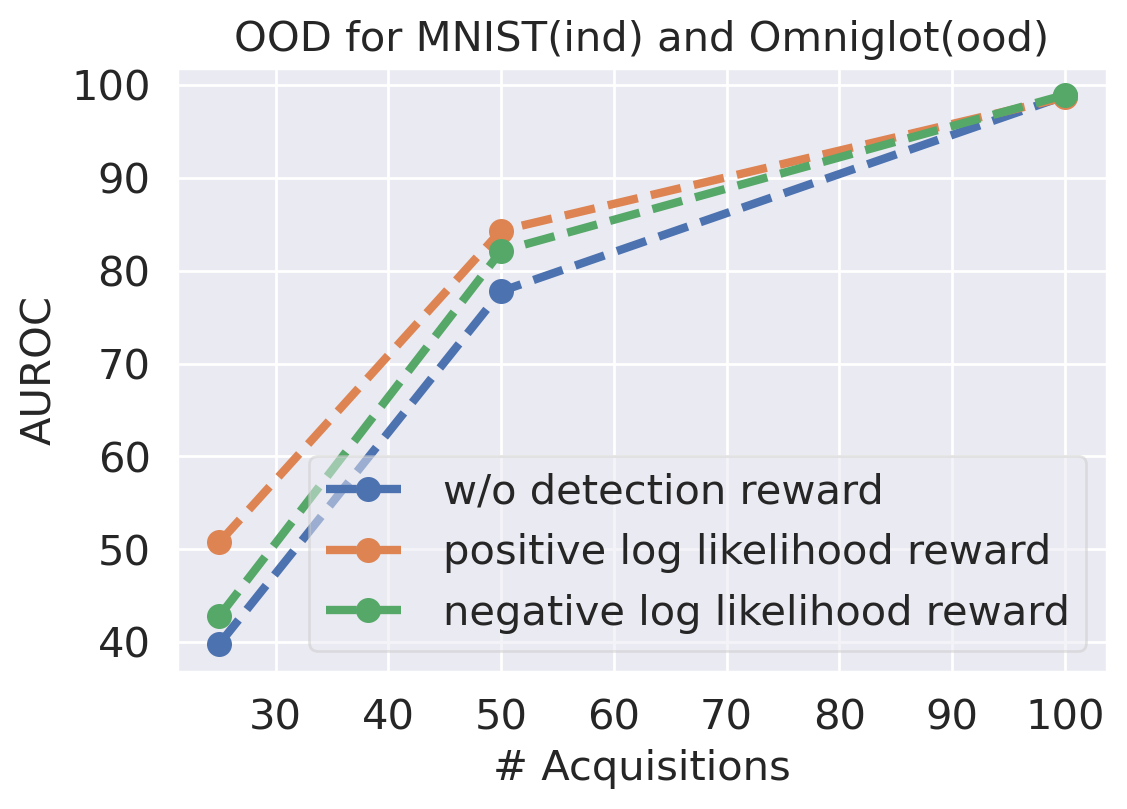}
    \caption{Comparison of different detection reward.}
    \label{fig:ablation_reward}
\end{figure}

\paragraph{Detection Reward} In the main text, we use $\log p(s_1,\ldots,s_{10} \mid \I_m)$ as an auxiliary reward from the OOD detector. As we discussed in Sec.~\ref{sec:robust_afa}, the positive log likelihood reward helps to reduce the false positive, while the negative log likelihood reward helps to reduce the false negative. Figure \ref{fig:ablation_reward} compares different types of detection reward. We can see both positive and negative likelihood reward can improve the detection performance, and the classification accuracy does not degrade a lot from the baseline.

\end{document}